\documentclass[runningheads]{llncs}

\usepackage{eccvabbrv}     
\usepackage{wrapfig}
\usepackage{colortbl}
\usepackage{xcolor}
\usepackage{float}
\usepackage{graphicx}
\usepackage{booktabs}
\usepackage{makecell}
\usepackage[table]{xcolor}

\usepackage{caption}
\usepackage[accsupp]{axessibility}  

\usepackage{hyperref}

\usepackage{orcidlink}

\begin{document}


\title{MonoMSK: Monocular 3D Musculoskeletal Dynamics Estimation}

\titlerunning{MonoMSK}

\author{Farnoosh Koleini \and
Hongfei Xue \and
Ahmed Helmy \and
Pu Wang}

\authorrunning{F.~Koleini et al.}

\institute{
University of North Carolina at Charlotte\\
\email{\{fkoleini, hongfei.xue, ahelmy1, Pu.Wang\}@charlotte.edu}
}

\maketitle


\begin{figure}[ht]
    \centering
    \includegraphics[width=0.85\linewidth]{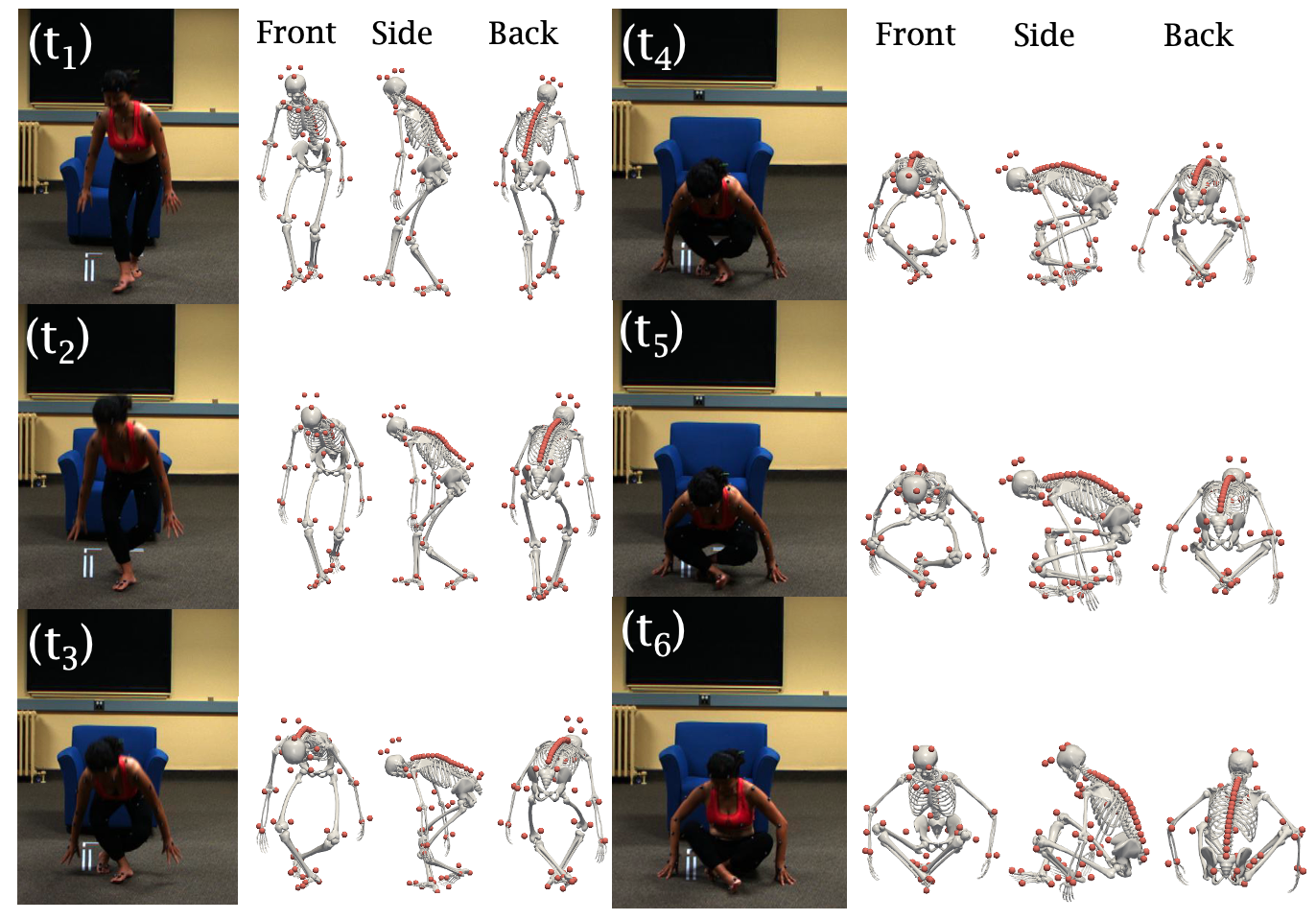}
    \captionsetup{width=0.85\linewidth}
    \caption{\textbf{MonoMSK} is a framework for physically grounded 3D human motion estimation from monocular videos.}
    \label{fig:landing}
    \vspace{-30pt}
\end{figure}


\begin{abstract}
Reconstructing biomechanically realistic 3D human motion,
recovering both kinematics (motion) and kinetics (forces), is a critical challenge.
While marker-based systems are lab-bound and slow, popular monocular methods use oversimplified,
anatomically-inaccurate models (e.g., SMPL) and ignore physics, fundamentally limiting their biomechanical fidelity.
In this work, we introduce MonoMSK, a hybrid framework that bridges data-driven learning and physics-based simulation
for biomechanically realistic 3D human motion estimation from monocular video.
MonoMSK jointly recovers both kinematics (motions) and kinetics (forces and torques) through an anatomically accurate
musculoskeletal model.
By integrating transformer-based inverse dynamics with differentiable forward kinematics and dynamics layers governed
by ODE-based simulation, MonoMSK establishes a physics-regulated inverse--forward loop that enforces biomechanical
causality and physical plausibility.
A novel forward--inverse consistency loss further aligns motion reconstruction with the underlying kinetic reasoning.
Experiments on BML-MoVi, BEDLAM, and OpenCap show MonoMSK significantly outperforms state-of-the-art methods in kinematic
accuracy, while, for the first time, enabling precise monocular kinetics estimations.

\end{abstract}

\section{Introduction}
\vspace{-5pt}
Understanding and reconstructing human motion with biomechanical and physical realism is a long-standing goal in computer vision, biomechanics, and robotics.
Biomechanically-accurate 3D human motion estimation aims to recover not only anatomically valid joint configurations but also the underlying physical quantities, such as forces and torques, that drive those motions.
Such physically-grounded representations are critical for applications in clinical motion analysis, sports science, rehabilitation, and human–robot interaction \cite{uhlrich2023opencap,seth2018opensim}. While Biomechanically-accurate motion estimation offers numerous benefits, its 
broad adoption is hindered by significant barriers. 

The gold standard approach for biomechanics-accurate motion estimation combines multi-camera, marker-based motion capture with biomechanical optimization tools, which are expensive,
labor-intensive, and time-consuming \cite{ceseracciu2014comparison}. In practice, current workflows use {multiple cameras to track the markers on the body \cite{westlund2015motion}. The captured marker data are then processed with time-consuming optimization-based simulations  such as OpenSim \cite{seth2018opensim}.  For example, the state-of-the-art OpenCap system takes approximately 2 minutes for kinematics (e.g., joint rotation angles and velocities) and 35 minutes for kinetics (e.g., joint force and torques)\cite{uhlrich2023opencap}.
Learning-based monocular motion estimation frameworks \cite{goel2023humans,dwivedi2024tokenhmr,patel2025camerahmr} have achieved impressive results in 3D human pose reconstruction by employing deep learning models to recover parametric body representations such as SMPL \cite{loper2023smpl} from single-camera monocular videos. Despite their strong visual realism and accessibility, these models typically rely on oversimplified skeletal structures with anatomically inaccurate joint positions and bone orientations, fundamentally limiting their biomechanical fidelity. Recent efforts have sought to mitigate these issues by incorporating biomechanically accurate skeletal models \cite{koleini2025biopose,jiang2024manikin}. However, they overlook the underlying physical laws that govern the causal relationships between forces (kinetics) and motions (kinematics). Consequently, these approaches are incapable of estimating kinetics and often exhibit reduced accuracy in kinematic reconstruction, as well.

To address these challenges, we introduce MonoMSK, a hybrid framework that integrates learning-based inverse dynamics with a differentiable, anatomically-accurate musculoskeletal (MSK) forward simulation to recover biomechanically-accurate motion dynamics from monocular video. MonoMSK comprises five integrated components: (1) a pretrained human-mesh-recovery model to extract anatomically-grounded virtual markers from the input video; (2) an Inverse Kinematics Transformer (IKT) to predict joint angles from these markers; (3) an Inverse Dynamics Transformer (IDT) to infer kinetic quantities (e.g., joint torques, ground-reaction forces) from the predicted kinematics; (4) a differentiable Forward Kinematics (FK) layer; and (5) a differentiable Forward Dynamics (FD) layer that leverages an ODE solver to simulate physically-consistent trajectories from the inferred kinetics. 

This novel integrated kinematics–kinetics design tightly couples data-driven temporal modeling with continuous-time physical simulation. Specifically, our data-driven inverse transformers (IKT, IDT) are trained to infer the kinetic causes (joint torques) from the observed kinematic consequences (joint motion). The differentiable FD and FK layers then act as a physics-based verifier: it uses these inferred kinetics to simulate the resulting motion, ensuring the predicted forces can faithfully reconstruct the original observation. Moreover, by leveraging anatomy-constrained FK and FD layers, this physics-regulated loop embeds domain knowledge during both training and inference, ensuring  estimated motion remains physically plausible and anatomically coherent.

\noindent Our key contributions are summarized as follows:
\begin{itemize}
\item We introduce MonoMSK, a hybrid motion dynamics estimation framework that couples learning-based inverse dynamics with a differentiable, anatomically accurate musculoskeletal forward simulator to recover physically grounded 3D motion dynamics (kinematics and kinetics) from monocular video
\item We introduce a biomechanics-informed training scheme, which combines optimal-control biomechanics simulations for ground-truth generations with  kinetics and kinematics supervision losses and forward-inverse consistency loss that  align the bidirectional translation between kinematics (e.g., motion) and kinetics (e.g., force). 

\item Experiments on BML-MoVi, BEDLAM, and OpenCap show MonoMSK significantly outperforms state-of-the-art methods in kinematic accuracy, while, for the first time, enabling precise monocular kinetics estimations. 
\end{itemize}
\vspace{-15pt}
\section{Related Work}
\label{sec:related work}
\vspace{-5pt}
\subsection{Monocular 3D Human Pose Estimation}
Monocular 3D human pose estimation has progressed rapidly with the advent of deep learning and parametric body models such as SMPL \cite{loper2023smpl}.
Early approaches \cite{kolotouros2019convolutional} employ convolutional networks to regress SMPL parameters from single images, while later works extend this to video sequences using temporal transformers \cite{zhang2024physpt}.
The introduction of transformer-based models such as HMR2.0 \cite{goel2023humans}, TokenHMR \cite{dwivedi2024tokenhmr}, CameraHMR \cite{patel2025camerahmr} and PhysPT \cite{zhang2024physpt} have significantly improved human mesh reconstruction accuracy.
However, these models are fundamentally limited for the applications that require biomechanical-accurate kinematics estimations. In particular, these models typically rely on oversimplified skeletal structures with anatomically inaccurate joint positions and bone orientations, fundamentally limiting their biomechanical fidelity. To address these limitations, we leverage the 3D human mesh reconstructed from the HMR models to extract virtual tracking markers, which then are leveraged by our MonoMSK model to infer physically grounded kinetics and kinematics estimations based on anatomically-accurate human MSK model. As a result,  MonoMSK can serve as the plug-and-play module for any HMR models so that they can produce biomechanically-accurate.

\vspace{-15pt}
\subsection{Biomechnically-accurate Motion Estimation}

The current gold standard for biomechanics analysis combines multi-camera, marker-based motion capture systems with optimization-based musculoskeletal solvers such as OpenSim~\cite{seth2018opensim}. 
These systems rely on retro-reflective markers tracked by synchronized infrared cameras in controlled laboratory environments. 
The recorded marker trajectories are then processed through inverse kinematics and dynamics pipelines to estimate joint torques, ground reaction forces, and muscle activations. 
Although these workflows achieve high biomechanical accuracy, they are expensive, labor-intensive, and limited to specialized lab setups. 
They also require expert supervision, precise calibration, and time-consuming optimal control simulations. 
For example, OpenCap~\cite{uhlrich2023opencap} reports approximately two minutes per trial for kinematic reconstruction and up to thirty-five minutes for kinetic estimation. Such constraints make these methods impractical for large-scale or real-world deployment. Recent work, such as BioPose, D3KE and HSMR has begun to incorporate biomechanically accurate skeletal models \cite{koleini2025biopose,bittner2022towards,xia2025reconstructing}. However, these methods typically omit explicit physical laws that govern the causal link between forces (kinetics) and motion (kinematics). As a result, they cannot estimate kinetic variables and often yield diminished accuracy in kinematic reconstruction. 

In contrast to prior approaches, our proposed \emph{MonoMSK} framework unifies kinematic perception and dynamic reasoning within a single, end-to-end trainable model. MonoMSK is a biomechanics/physics refinement module, not an HMR backbone. It uses SMPL mesh vertices as virtual markers with anatomy-constrained kinematic alignment and dynamics-based supervision, yielding improved joint accuracy and MPBLPE even from identical HMR
mesh inputs. Beyond BioPose’s kinematics-only pipeline \cite{koleini2025biopose}, MonoMSK also estimates joint torques and GRFs and enforces temporal physical consistency via differentiable forward dynamics (ODE).
By embedding a differentiable physics solver directly into a transformer-based motion prediction architecture, MonoMSK simultaneously learns inverse dynamics (force–torque inference) and forward dynamics (motion generation) without separate optimization stages. 
This integration enables physically stable, temporally coherent, and anatomically consistent motion trajectories—achieving biomechanical accuracy comparable to laboratory systems while maintaining the scalability and flexibility of modern deep learning. 
\vspace{-25pt}
\section{Proposed Method: MonoMSK}
\label{sec:proposed method}
\begin{figure*}[htbp]
    \centering
    \includegraphics[width=1.0\linewidth]{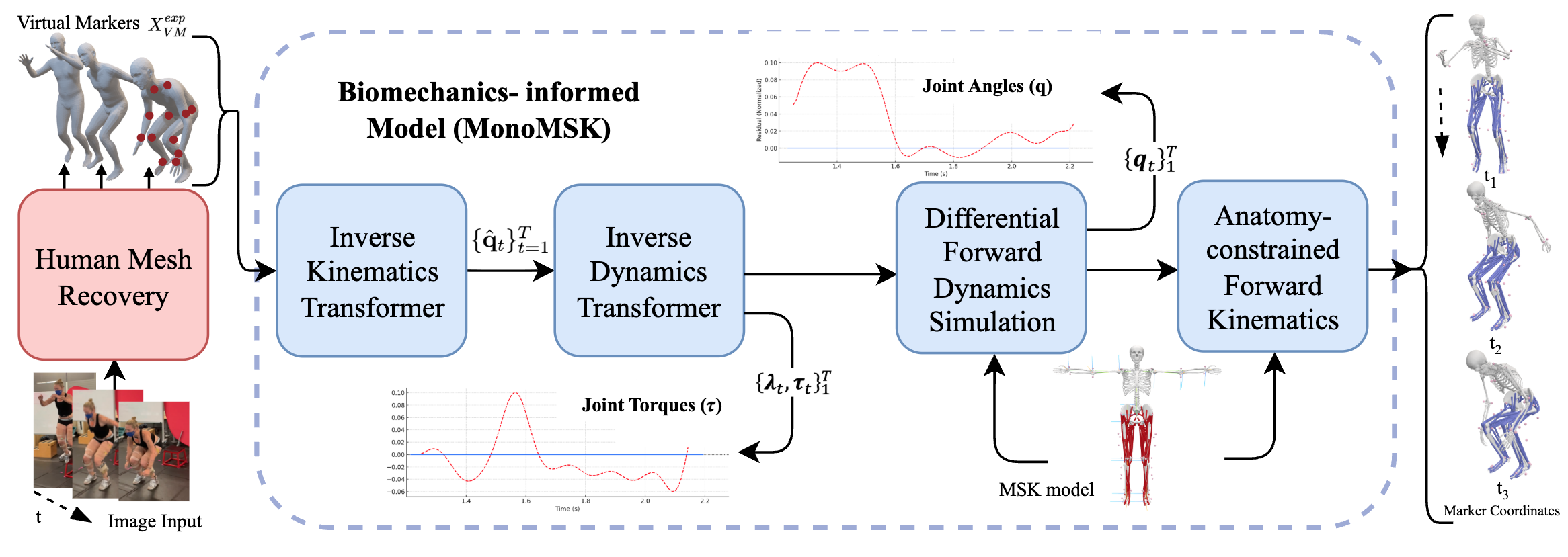}
\caption{
Overview of the MonoMSK pipeline. A monocular video is processed by a pretrained Human Mesh Recovery (HMR) model to obtain 3D meshes and virtual markers. The \textbf{Inverse Kinematics Transformer (IKT)} converts these markers into anatomically constrained musculoskeletal joint states $\mathbf{q}$. The \textbf{Inverse Dynamics Transformer (IDT)} infers the latent dynamic quantities, internal torques $\boldsymbol{\tau}$ and external ground-reaction forces $\boldsymbol{\lambda}$. A differentiable \textbf{Forward Dynamics (FD)- ODE solver} ODE solver integrates these forces through the Euler–Lagrange MSK dynamics to produce physically coherent future motion.
}
    \label{fig:MonoMSK_overview}
\end{figure*}

\vspace{-5pt}
The objective of MonoMSK is to estimate biomechanically accurate and physically consistent 3D human motion directly from monocular video. As illustrated in Figure~\ref{fig:MonoMSK_overview}, MonoMSK builds upon a detailed musculoskeletal (MSK) model and integrates kinematic inference with physics-based dynamic reasoning. The pipeline begins with a pretrained Human Mesh Recovery (HMR) models, which regress pose and shape parameters and generate SMPL meshes whose vertices serve as virtual motion-capture markers (\S\ref{sec:biomech-estimation}). These marker trajectories are processed by an Inverse Kinematics Transformer (IKT) (\S\ref{sec:IKT}) to estimate the generalized MSK kinematic state $\mathbf{q}=\{\mathbf{T},\mathbf{R},\mathbf{q}^{r}\}$, capturing anatomically valid joint rotations. The recovered kinematics are then provided to the Inverse Dynamics Transformer (IDT) (\S\ref{sec:IDT}), which predicts internal joint torques $\boldsymbol{\tau}$ and external ground-reaction forces $\boldsymbol{\lambda}$ that generate the observed motion. To enforce biomechanical correctness, MonoMSK integrates a differentiable Forward Dynamics (FD) layer based on an ODE formulation of the MSK dynamics (\S\ref{sec:ODE}). This layer simulates the forward-time evolution of the body under the predicted forces and torques, producing physically coherent kinematic states that are matched back to the IKT estimates through inverse–forward consistency. Together, these components form a unified, end-to-end differentiable framework that tightly couples perception with musculoskeletal physics, enabling accurate, stable, and interpretable human motion reconstruction.
\vspace{-10pt}
\subsection{Human Musculoskeletal Model}\label{sec:bsk}
The biomechanical human model consists of two
core components: the musculoskeletal (MSK) body model and the musculoskeletal (MSK) dynamics model (See Figure~\ref{fig:MSk}).

\noindent\textbf{MSK Body  Model.}
The MSK body model, e.g., the widely-adopted OpenSim model, represents the body as $N_s=24$ rigid bone segments interconnected by anatomically constrained joints. Based on LaiUhlrich2022 \cite{uhlrich2023opencap}, our MSK model features a flexible spine and full shoulder mobility. It includes a 3-DoF trunk joint (lumbar extension, lateral bending, and axial rotation) and 3-DoF shoulder joints on both sides (arm flexion/extension, abduction/adduction, and internal/external rotation). Each joint $i$ defines motion according to its physiological Degrees of Freedom (DoFs) $D_i$, thus determining the human body’s valid kinematic and dynamic configuration space. Each joint $i$ is parameterized by the anatomy-dependent bone orientation $q_i^{o}\!\in\!\mathbb{R}^{3}$ (Appendix Sec.~\ref{sec:MSK_SMPL}) , muscle-induced joint rotation $q_i^{r}\!\in\!\mathbb{R}^{D_i}$ ($D_i\!\leq\!3$), and bone scaling $s_i\!\in\!\mathbb{R}^{3}$~\cite{featherstone2008rigid,murray2017mathematical}. 
\begin{wrapfigure}{r}{0.35\textwidth}  
    \vspace{-10pt}
    \centering
    \includegraphics[width=0.40\textwidth]{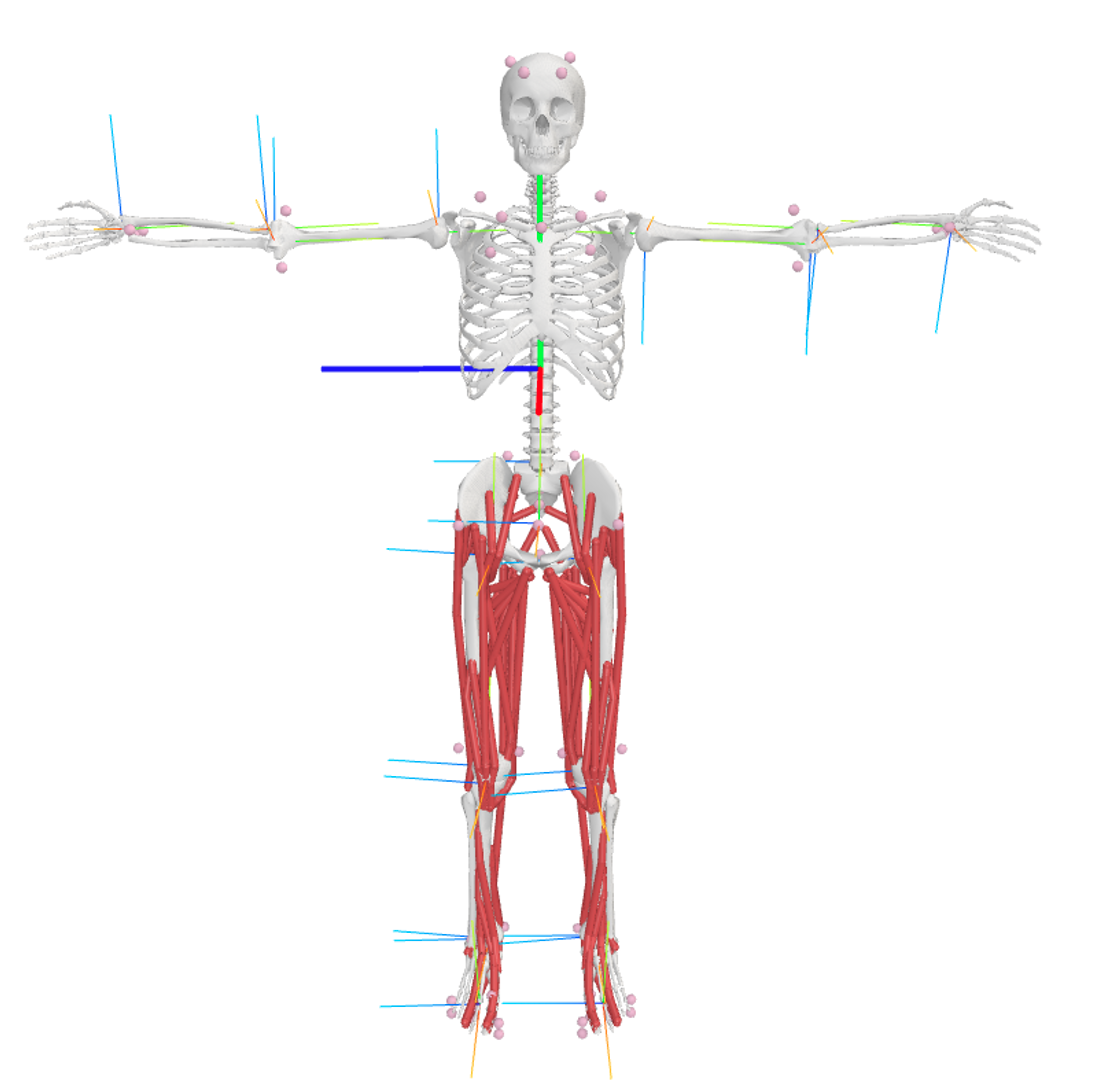}
    \caption{Musculoskeletal (MSK) body model with anatomically precise joint positions, bone orientations, and muscle geometry (red). Pink spheres indicate virtual model markers attached to bone segments for accurate biomechanical tracking.}
    \label{fig:MSk}
    \vspace{-10pt}
\end{wrapfigure}
The scaling process tailors a generic anatomical MSK body model to fit subject-specific body geometry, such as bone lengths and muscle attachment points. (Appendix Sec.~\ref{sec:MSK_Model_Scaling}). These parameters  $(\mathbf{q}^{o}, \mathbf{q}^{r}, s)$ collectively define the entire skeleton ~\cite{delp2007opensim}. In MSK model, each joint transform is defined by a fixed, anatomy-dependent joint frame/orientation  $q^o$ and a time-varying joint coordinate \(q^r\) (the physiological DoFs, e.g., flexion/extension). Eq.~(1) defines the generalized coordinates \(q\) as these \emph{time-varying} joint coordinates (plus global pose/translation), while the anatomy-dependent bone/joint orientations are \emph{not} part of \(q\) and remain constant during inference. The 3D motion kinematics can be described compactly in a generalized coordinate system as
\begin{equation}
\mathbf{q} = \{\mathbf{T}, \mathbf{R}, \mathbf{q}^{r}\},
\label{eq:generalized_coordinate}
\end{equation}
where $\mathbf{T}\!\in\!\mathbb{R}^3$ and $\mathbf{R}\!\in\!\mathbb{R}^3$ represent global translation and rotation of the root segment, and $\mathbf{q}^{r}$ encodes the motion-induced joint rotations of the 24-joint MSK model. Correspondingly, $\dot{\mathbf{q}}$ and $\ddot{\mathbf{q}}$ are used to represent the velocity and acceleration of generalized coordinates (Appendix Sec.~\ref{sec:MSK_SMPL}).

\noindent \textbf{Muscle-driven Forward Dynamics.} It uses differential-algebraic equations to predict body motion from muscle excitations.  Muscles attach to skeleton bones through defined paths to determine fiber and tendon lengths, which, combining with body segment parameters (e.g., mass, center of mass, and inertia), generates the forces and torques applied at joints, according to muscle activation, contraction dynamics, and multibody dynamics \cite{caruel2018physics}. In particular, multibody dynamics, following the Newton-Euler equations of motion, defines forward dynamics that predict motion from internal joint torques $\boldsymbol{\tau}$ and external forces $\boldsymbol{\lambda}$ (e.g., ground reaction forces, GRFs) 
\begin{equation}
\mathbf{J}_C^{\mathsf{T}}\boldsymbol{\lambda} + \boldsymbol{\tau} = \mathbf{M}(\mathbf{q})\ddot{\mathbf{q}} + \mathbf{C}(\mathbf{q},\dot{\mathbf{q}}) + \mathbf{g}(\mathbf{q}),
\label{eq:euler_lagrange}
\end{equation}
where $\mathbf{M}$, $\mathbf{C}$, and $\mathbf{g}$ denote the generalized inertia matrix, Coriolis/centrifugal terms, and gravitational generalized torques, respectively. $\mathbf{J}_C$ is the contact Jacobian matrix that translates the generalized velocities $\dot{\mathbf{q}}$ to the velocities at the point of contact between foot and ground. Each joint torque $\tau_i$ is expressed as the net contribution of muscle forces transmitted through moment arms:
\begin{equation}
\tau_i = \sum_{j=1}^{N_m} r_{ij}\, F_j(a_j, l_j, v_j) + \tau_i^{\mathrm{tm}},
\label{eq:muscle_torque}
\end{equation}
where $r_{ij}$ is the moment arm of muscle $j$ about joint $i$, $N_m$ is the number of muscles, $a_j$ is the muscle activation, $l_j$ and $v_j$ are fiber length and velocity, and $\tau_i^{\mathrm{tm}}$ denotes an ideal torque motor at major joints (e.g., lumbar, shoulder, elbow) for residual actuation. Muscle forces $F_j$ follow a Hill-type model \cite{caruel2018physics}: $F_j(a_j, l_j, v_j) = a_j F_{j}^{\max} f_l(l_j) f_v(v_j) + F_{j}^{\mathrm{pass}}(l_j),$
where $F_{j}^{\max}$ is maximal isometric force, $f_l$ and $f_v$ are the force–length and force–velocity relationships, and $F_{j}^{\mathrm{pass}}$ represents passive elastic tension.  


\noindent\textbf{Ground Reaction Forces.}
To simulate realistic foot-ground interactions, ground reaction forces are modeled through six foot-ground contact spheres attached to the foot segments (Appendix Sec.~\ref{sec:Contact}). Each contact sphere $k$ generates normal and tangential ground reaction forces modeled by the compliant Hunt–Crossley formulation~\cite{hunt1975coefficient}:
\begin{equation*}
\mathbf{F}_{n,k} = k_n \delta_k^{1.5} (1 + c_n \dot{\delta}_k)\mathbf{n}, \quad
\mathbf{F}_{t,k} = -\mu \|\mathbf{F}_{n,k}\| \frac{\mathbf{v}_{t,k}}{\|\mathbf{v}_{t,k}\| + \epsilon},
\label{eq:contact}
\end{equation*}
where $\delta_k$ and $\dot{\delta}_k$ are the penetration depth and velocity, $\mathbf{n}$ the ground normal, $\mathbf{v}_{t,k}$ the tangential velocity, and $(k_n, c_n, \mu)$ are stiffness, damping, and friction coefficients. The total ground reaction force is
\begin{equation}
\boldsymbol{\lambda} = \sum_k (\mathbf{F}_{n,k} + \mathbf{F}_{t,k}),
\label{eq:grf}
\end{equation}

\noindent This physically grounded contact model captures smooth transitions through heel-strike, mid-stance, and toe-off phases~\cite{rajagopal2016full,seth2018opensim}.
\vspace{-15PT}
\subsection{Biomechanics-informed Motion Estimation Model (MonoMSK)} 
\label{sec:biomech-estimation}
Built on top of the MSK model, we introduce MonoMSK to estimate kinematics and kinetics from monocular videos. First, we will leverage the existing monocular human mesh recovery (HMR) models, which estimate 3D pose $\theta$ and shape $\beta$ parameters to generate a 3D mesh via a parametric model like SMPL. Any subset of the 6,890 vertices on the SMPL mesh can act as virtual motion capture markers. Leveraging these virtual marker tracking data as inputs, MonoMSK will first learn to estimate kinematics 
$\mathbf{q} = \{\mathbf{T}, \mathbf{R}, \mathbf{q}^{r}\}$ via Inverse Kinematics Transformer (IKT). Then, the kinematic estimation $\mathbf{q}$ serves as the input of the Inverse Dynamics Transformer (IDT) to predict kinetic attributes, including internal joint torques $\boldsymbol{\tau}$ and external contact forces $\boldsymbol{\lambda}$.  To incorporate biomechanics priors into the network architecture, we directly inject a physics-based ODE solver into the network, which leverages forward dynamics to transform the estimated joint torques and contact forces $\{\boldsymbol{\tau}, \boldsymbol{\lambda} \}$  back to joint kinematics $\mathbf{q}$. The kinematic estimation $\mathbf{q}$ will then be translated into joint positions via an anatomical-aware forward kinematic layer. Inverse-forward consistency training is leveraged to ensure the biomechanically accurate motion estimations. 
\vspace{-10pt}
\subsubsection{Inverse Kinematics Transformer (IKT)}
\label{sec:IKT}
Given a monocular RGB video sequence $\{I_t\}_{t=1}^T$, we employ a pretrained human mesh recovery model to regress per-frame pose and shape parameters $\{\hat{\boldsymbol{\theta}}_t, \hat{\boldsymbol{\beta}}_t\}_{t=1}^T$. The pose $\boldsymbol{\theta}_t \in \mathbb{R}^{23\times3}$ encodes axis-angle rotations of the 23 joints, while the shape $\boldsymbol{\beta}_t \in \mathbb{R}^{10}$ represents low-dimensional body morphology \cite{loper2023smpl}. By combining these pose and shape parameters, the SMPL model produces a detailed 3D mesh \(M(\theta, \beta) \in \mathbb{R}^{3 \times N}\), where \(N = 6890\) vertices represent the surface of the body. The positions of the body joints \(J \in \mathbb{R}^{3 \times k}\) are then derived as a weighted sum of these vertices, formulated as \(J = MW\), where \(W \in \mathbb{R}^{N \times 23}\) contains the predefined linear blending weights that map vertices to the joints. With the joint positions $J$ as inputs, we employ the pretrained global trajectory predictor~\cite{carreira2016human} to estimate per-frame translation and rotation $\{\hat{\mathbf{T}}_t, \hat{\mathbf{R}}_t\}_{t=1}^T$ (Appendix Sec.~\ref{sec:global_traj}).

Since the SMPL model employs a simplified skeleton rig with inaccurate joint location and bone orientations, $\boldsymbol{\theta}_t \in \mathbb{R}^{23\times3}$ cannot be directly used to estimate anatomical joint rotation $\mathbf{q}^{r} \in \mathbb{R}^{24\times3}$. Following BioPose~\cite{koleini2025biopose},  we select a set of $M$ tracking markers $\mathbf{x} = \{x^{i}\}_{i=1}^M$  that attached to the bone segments in such a way that each bone segment is associated with at least $D_i$ markers to ensure the unique solutions of the derived rotation angles at joint $i$ with $D_i$ degrees of freedom.  The transformer encoder takes a  sequence of marker data $\{\mathbf{x}_t\}_{t=1}^T$ over $T$ frames as inputs to predict the muscle-induced joint rotation $\{\hat{\mathbf{q}_t^{r}}\}_{t=1}^T= \mathrm{E}_{\text{IKT}}(\{\mathbf{x_t}\}_{t=1}^T)$ which, combing with predicted global translation and rotation, yields the kinematic estimation ~\cite{zhang2024physpt}
\[
\{\hat{\mathbf{q}}_t\}_{t=1}^T = \{\hat{\mathbf{T}}_t, \hat{\mathbf{R}}_t, \hat{\mathbf{q}^{r}}_t \}_{t=1}^T
\]

\subsubsection{Inverse Dynamics Transformer (IDT)}
\label{sec:IDT}
Given the kinematic estimations $\{\hat{\mathbf{q}}_t\}_{t=1}^T$ from IKT encoder, the IDT predicts the joint torques and contact forces  $\{\boldsymbol{\tau}_t, \boldsymbol{\lambda}_t \}_{t = 1}^T$. At each time step $t$, a motion feature vector $\mathbf{f}_t \in \mathbb{R}^{f_{\text{dim}}}$ is formed by concatenating the generalized coordinates and their temporal derivatives, i.e., $\mathbf{f}_t = [\hat{\mathbf{q}}_t, \hat{\dot{\mathbf{q}}}_t]$. 
Each feature is projected into the transformer embedding space using an MLP encoder:
\[
\mathbf{z}_t = \mathrm{MLP}_{\text{enc}}(\mathbf{f}_t) \in \mathbb{R}^{d_{\text{model}}},
\]
followed by sinusoidal positional encoding $\mathrm{PE}(\cdot)$ to maintain temporal order. 
The encoded sequence $\mathbf{z}_{1:T}$ is then processed by a transformer encoder:
\[
\mathbf{h}_{1:T} = \mathrm{E}_{\text{IDT}}(\mathrm{PE}(\mathbf{z}_{1:T})),
\]
yielding temporally contextualized motion embeddings $\mathbf{h}_{1:T}$ that capture biomechanical dependencies and long-range temporal context across the motion sequence. From $\mathbf{h}_{1:T}$, a dedicated MLP head predicts the per-frame kinetics parameters:
\[
\{\hat{\boldsymbol{\lambda}}_t, \hat{\boldsymbol{\tau}}_t\}_{t = 1}^{T} = \mathrm{MLP}_{\text{force}}(\mathbf{h}_{1:T}),
\]



\subsubsection{Forward Dynamics (FD) Layer via ODE Solver}
\label{sec:ODE}
Given the predicted kinetics $\{\hat{\boldsymbol{\lambda}}_t, \hat{\boldsymbol{\tau}}_t\}_{t = 1}^{T}$ from IDT, the FD layer predict continuous-time kinematics via the differential ODE solver, which numerically integrates the human body’s motion over time using the the Newton-Euler equations of motions defined in Eq. \eqref{eq:euler_lagrange}, enabling simulation of future body states consistent with physical laws. The kinematic state at time $t$ is
\[
\mathbf{x}_t = [\,\mathbf{q}_t, \dot{\mathbf{q}}_t\,]^{\mathsf{T}},
\]
where $\mathbf{q}_t=\{\mathbf{T}_t,\mathbf{R}_t,\boldsymbol{q}_t^r\}$ represents generalized states and $\dot{\mathbf{q}}_t$ their velocities. 
The second-order dynamics are rewritten as a first-order ODE system $f_{\mathrm{ODE}}(\mathbf{x}_t)$:
\begin{equation}
\frac{d}{dt}
\begin{bmatrix}
\mathbf{q}_t \\ \dot{\mathbf{q}}_t
\end{bmatrix}
=
\begin{bmatrix}
\dot{\mathbf{q}}_t \\ 
\mathbf{M}^{-1}(\mathbf{q}_t)
\!\left(\mathbf{J}_C^{\mathsf{T}}\boldsymbol{\lambda}_t
+ \boldsymbol{\tau}_t
- \mathbf{C}(\mathbf{q}_t,\dot{\mathbf{q}}_t)
- \mathbf{g}(\mathbf{q}_t)\right)
\end{bmatrix}\!,
\label{eq:ode_small}
\end{equation}
which defines a differentiable mapping that governs the temporal evolution of the biomechanical system. To compute the predicted trajectory $\{\hat{\mathbf{q}}_{t+n}\}_{n=1}^{N}$, we integrate Eq.~\eqref{eq:ode_small} forward in time using a differentiable ODE solver. 
In practice, a fourth-order Runge–Kutta (RK4) or adaptive Dormand–Prince (RK45) solver can be adopted~\cite{zhang2024incorporating}, which ensures numerical stability while maintaining differentiability for backpropagation:
\begin{equation}
\mathbf{x}_{t+\Delta t} = \mathrm{ODE_{solver}}\big(f_{\mathrm{ODE}}, \mathbf{x}_t, \Delta t;\, \boldsymbol{\phi}\big),
\label{eq:ode_solver}
\end{equation}
where $\boldsymbol{\phi}$ denotes physical parameters (e.g., segmental masses, inertia, joint stiffness, and damping). 
The solver predicts subsequent generalized states as:
\begin{equation}
\begin{aligned}
\dot{\mathbf{q}_{t+1}}^{fd} &= \dot{\mathbf{q}_{t}}^{fd} + \ddot{\mathbf{q}_{t}}^{fd}\Delta t, \mathbf{q}_{t+1}^{fd} &= \mathbf{q}_{t}^{fd} + \dot{\mathbf{q}_{t}}^{fd}\Delta t,
\end{aligned}
\label{eq:forward_update}
\end{equation}
This iterative process simulates motion under the learned forces and torques, producing continuous, dynamically coherent trajectories.


\subsubsection{Anatomical Forward Kinematic (FK) Layer}
The FK layer transforms the rest-pose anatomical joint markers to the new positions according to the ODE-simulated kinematic dynamics $\{{\mathbf{q}}^{fd}_t\}_{t=1}^T$. In particular, the global position of joint $i$ is computed recursively through:
\begin{equation}
\mathbf{J}_i^{fk} = \mathbf{R}_{i}(\mathbf{J}_i^{0} \odot s_i) + \mathbf{J}_{\mathrm{par}(i)},
\label{eq:bsk_fk}
\end{equation}
where $\mathbf{J}_i^{0}$ is the local joint position offset, $\odot$ denotes element-wise bone scaling, and $\mathrm{par}(i)$ indicates the parent joint. The corresponding rotation is obtained as $\mathbf{R}_i = \mathbf{R}_{\mathrm{par}(i)}\, \mathbf{R}(q_i^{o})\, \mathbf{R}(q_i^{r})$. Since we employed a full-body biomechanical MSK model, the FK transformation is inherently constrained by the realistic joint degrees of freedom  $q_i^{r}\!\in\!\mathbb{R}^{D_i}$ ($D_i\!\leq\!3$), bone orientations $q^{o}$ and scales $s_i$.  
\vspace{-10pt}
\subsection{Biomechanics-informed Model Training} 
To enforce biomechanics-accurate estimations, our model is trained with multiple physics-informed losses including supervision losses on kinematic states $\{\hat{\mathbf{q}}_t\}_{t=1}^T$  and kinetic forces $\{\hat{\boldsymbol{\lambda}}_t, \hat{\boldsymbol{\tau}}_t\}_{t = 1}^{T}$  along with the consistency loss between inverse and forward motion dynamics.  

\noindent\textbf{Ground-truth Dynamics via Optimal-control Forward Simulation.} To facilitate supervised training, we obtain ground-truth MSK dynamics via optimal control simulations in physics engines like OpenSim-Moco \cite{dembia2020opensim}. This has become the gold standard practice for non-invasive measurement of variables such as muscle forces and joint torques, which are difficult to measure directly in vivo. In particular, the goal is to find the optimal muscle excitation signals $\mathbf{e}$ that minimize muscular effort and maximize agreement between simulated and observed motion under the constraints of  Newton-Euler equations of motion defined in Eq.\eqref{eq:euler_lagrange}, i.e., 
\begin{equation*}
\begin{aligned}
\arg\min_{e} \int_{t_0}^{t_f} \Big(
  & w_1 \|\mathbf{e}_{t}\|_2^2 
  + w_2 \|\mathbf{q}_t - \tilde{\mathbf{q}}_t\|_2^2
  + w_3 \|\dot{\mathbf{q}}_t - \tilde{\dot{\mathbf{q}}}_t\|_2^2
  + w_4 \|\ddot{\mathbf{q}}_t - \tilde{\ddot{\mathbf{q}}}_t\|_2^2
\Big)\, dt,
\end{aligned}
\label{eq:objective_function}
\end{equation*}
where $w_i$ are weighting coefficients and $(\tilde{\mathbf{q}}, \tilde{\dot{\mathbf{q}}}, \tilde{\ddot{\mathbf{q}}})$ are ground-truth kinematics derived from inverse kinematics optimization based on the marker tracking data detailed in the Supplementary material. The optimal control simulation can be transcribed into a finite-dimensional nonlinear program using direct collocation, which is then solved using a large-scale nonlinear optimization solver such as IPOPT~\cite{wachter2006implementation}. The simulation outputs are reference torque $\tilde{\mathbf{\tau}}_t$ and force $\tilde{\mathbf{\lambda}}_t$.  

\noindent\textbf{Kinetics Losses.}
By training the model with supervised losses for the contact force and joint
torque, it will learn the physiological dynamics of the MSK system. Based on the reference torque $\tilde{\mathbf{\tau}}_t$ and force $\tilde{\mathbf{\lambda}}_t$, the training objective is to minimize the mean squared absolute errors
\begin{equation}
\mathcal{L}_{\text{kinetic}}
= w_{\lambda} \, \mathcal{L}_{\lambda}
+ w_{\tau} \, \mathcal{L}_{\tau},
\label{eq:IDT_loss}
\end{equation}
  \vspace{-20pt}
where
\[
\mathcal{L}_{\lambda} = \sum_{t=1}^{T}
\|\hat{\boldsymbol{\lambda}}_t - \tilde{\boldsymbol{\lambda}}_t\|_2^2,
\qquad
\mathcal{L}_{\tau} = \sum_{t=1}^{T}
\|\hat{\boldsymbol{\tau}}_t - \tilde{\boldsymbol{\tau}}_t\|_2^2
\]
$\mathcal{L}_{\lambda}$ promotes accurate estimation of foot–ground contact, and $\mathcal{L}_{\tau}$ enforces the accurate joint torque estimation. MonoMSK currently assumes rigid, planar ground contact in its contact/ground model, so it is not expected to generalize to deformable or non-rigid surfaces. Also, due to the emphasis on foot-contact modeling, the framework may exhibit slightly reduced accuracy in upper-body motion prediction.

\noindent \textbf{Inverse-Forward Consistency Losses.}
To embed the biomechanics causality into the model, we exploit the property that the translation between kinematics (e.g., motion) and kinetics (e.g., force) should be consistent. In particular, if the model inversely translates from kinematic consequences (e.g., rotation angles and joint positions) to the kinetic causes (e.g., muscle forces and joint torque), it should be able to arrive back at the original kinematic observations from its own kinetics predictions through physics-governed forward reasoning (e.g., the forward dynamics ODE simulations). In particular, we adopt two consistency losses 
\begin{equation}
\mathcal{L}_{\text{con}} 
= w_{q}\mathcal{L}_{q} + w_{J}\mathcal{L}_{J},
\label{eq:ode_total_loss}
\end{equation}
  \vspace{-20pt}
where
\[
\mathcal{L}_{q} = \sum_{t=1}^{T}
\|\hat{\mathbf{q}}^{fd}_t - \tilde{\mathbf{q}}_t\|_2^2 
\qquad
\mathcal{L}_{J} = \sum_{t=1}^{N}
\|\hat{\mathbf{J}}^{fk}_t  - \tilde{\mathbf{J}}_t\|_2^2.
\]
Here, $\tilde{\mathbf{q}}_t$ and $ \tilde{\mathbf{J}}_t$ denote ground-truth joint rotations and positions, while $\hat{\mathbf{q}}^{fd}_t$ are joint rotations obtained from the forward dynamics ODE solver and   $\hat{\mathbf{J}}^{fk}_t$ are joint positions derived via anatomical-constrained forward kinematics. Moreover, the ODE-based forward kinetics module functions as a differentiable physics simulator embedded within the transformer’s decoder loop. During training, the gradients of $\mathcal{L}_{\text{ODE}}$ are backpropagated through both the neural network parameters and the ODE solver, allowing the model to learn intrinsic dynamic properties—such as stiffness, damping, and contact responses—directly from data. This integration tightly couples data-driven temporal modeling with continuous-time physics, resulting in biomechanically faithful and dynamically smooth human motion trajectories. Furthermore, through the physics-regulated and anatomy-constrained FD and FK layers, domain knowledge is embedded not only during training but also during inference, ensuring consistent physical plausibility and anatomical coherence throughout the generation process.
\vspace{-10pt}
\section{Experiments}
\label{sec: experiments}
\vspace{-10pt}
\textbf{Datasets.} During training, we use BML-MoVi ~\cite{ghorbani2021movi}, which provides rich biomechanical motion capture and video data from multiple actors performing everyday activities. BMLMovi consists of 90 subjects performing 21 different actions, captured using two cameras and a marker-based motion capture system. For evaluation, we utilize the test set of OpenCap and BEDLAM, MonoMSK is \textbf{\emph{not}} trained on OpenCap ~\cite{uhlrich2023opencap} and BEDLAM ~\cite{black2023bedlam} datasets to show its cross-dataset generalization performance. OpenCap includes data from ten subjects performing various actions such as walking, squatting, standing up from a chair, drop jumps, and their asymmetric variations. The recordings were made using five RGB cameras alongside a marker-based motion capture system. Additionally, OpenCap offers processed marker data and kinematic annotations for a comprehensive full-body OpenSim skeletal model ~\cite{seth2018opensim}.  
BEDLAM dataset comprises synthetic video data featuring a total of 271 subjects, including 109 men and 162 women. It includes monocular RGB videos, covering a diverse range of body shapes, and motions (Appendix Sec.~\ref{sec:Datasets}).

\noindent\textbf{Implementation.} 
The transformer backbone employs a standard encoder-decoder
 architecture consisting of 6 layers, 8 attention heads, and an embedding dimension of 1024. 
The model operates on input sequences of length 16, consistent with common configurations in prior works. 
For efficient training, we first use the ground-truth supervision to extract output embeddings for 20 epochs, followed by an additional 5 epochs using predicted embeddings. 
The Adam optimizer is utilized with a weight decay of $10^{-4}$. 
The initial learning rate is set to $10^{-5}$ and is decayed by a factor of 0.8 every 5 epochs. 
Empirically, the hyperparameters are configured as 
$\gamma_{q} = 2\times10^{3}$, 
$\gamma_{J} = 1\times10^{5}$, 
$\gamma_{\tau} = 5$, 
$\gamma_{\lambda} = 1$, 
$\gamma_{v} = 100$, 
and $\gamma_{z} = 200$.

\noindent\textbf{Evaluation Metrics.}
We evaluate MonoMSK using both kinematic and dynamic measures. 
Kinematic accuracy is assessed using the MPBLPE (mm), which measures 3D bony-landmark positional error, and MAE$_{\textit{angle}}$ (deg), which quantifies joint-angle accuracy. 
To evaluate physical plausibility, we report the acceleration error (ACCL) and velocity error (VEL), capturing deviations in joint accelerations and velocities that reflect dynamic consistency. 
Finally, for direct kinetics estimation, we compute MAE$_{\lambda}$ (ground-reaction forces) and MAE$_{\tau}$ (joint torques), which assess the correctness of predicted external and internal forces (Appendix Sec.~\ref{sec:Evaluation_Metrics}).

\vspace{-10pt}
\subsection{Comparison to State-of-the-art Approaches}

\begin{table*}[!t]
\centering
\caption{We evaluate mean per--bony--landmark position error (MPBLPE, mm), joint--angle mean absolute error (MAE$_{\textit{angle}}$, deg), and 
physics--plausibility metrics acceleration error (ACCL, mm/frame$^2$) and velocity error (VEL, mm/frame). 
Lower values ($\downarrow$) indicate better performance. 
\textcolor{blue}{Blue} percentages represent relative improvements over the best kinematics-only baseline (BioPose). 
MonoMSK uses the MQ-HMR in \cite{koleini2025biopose} to extract virtual markers.}
\vspace{3pt}
\resizebox{\textwidth}{!}{%
\begin{tabular}{l|cccc|cccc|cccc}
\toprule
 & \multicolumn{4}{c|}{\textbf{BML-MoVi}} 
 & \multicolumn{4}{c|}{\textbf{BEDLAM}} 
 & \multicolumn{4}{c}{\textbf{OpenCap}} \\
\textbf{Methods} &
\makecell{MPBLPE\\(↓)} & 
\makecell{MAE$_{\textit{angle}}$\\(↓)} &
\makecell{Acc.\\(↓)} &
\makecell{Vel.\\(↓)} &
\makecell{MPBLPE\\(↓)} &
\makecell{MAE$_{\textit{angle}}$\\(↓)} &
\makecell{Acc.\\(↓)} &
\makecell{Vel.\\(↓)} &
\makecell{MPBLPE\\(↓)} &
\makecell{MAE$_{\textit{angle}}$\\(↓)} &
\makecell{Acc.\\(↓)} &
\makecell{Vel.\\(↓)} \\
\midrule

HMR2.0 \cite{goel2023humans} &
48.32 & 3.78 & 8.70 & 6.28 &
53.21 & 3.92 & 9.58 & 6.92 &
50.16 & 3.81 & 9.03 & 6.52 \\

TokenHMR \cite{dwivedi2024tokenhmr} &
44.54 & 3.54 & 8.02 & 5.79 &
48.42 & 3.69 & 8.72 & 6.29 &
46.17 & 3.57 & 8.31 & 6.00 \\

CameraHMR \cite{patel2025camerahmr} &
39.63 & 3.28 & 7.13 & 5.15 &
42.17 & 3.39 & 7.59 & 5.48 &
39.48 & 3.31 & 7.11 & 5.13 \\

\midrule

D3KE \cite{bittner2022towards} &
36.98 & 3.54 & -- & -- &
39.45 & 6.72 & -- & -- &
38.62 & 5.92 & -- & -- \\

OpenCap Multi-Camera \cite{uhlrich2023opencap} &
-- & -- & -- & -- &
-- & -- & -- & -- &
-- & 4.50 & -- & -- \\

\midrule

BioPose \cite{koleini2025biopose} &
\underline{25.76} & \underline{2.84} & \underline{6.30} & \underline{5.18} &
\underline{26.54} & \underline{3.14} & \underline{6.84} & \underline{4.37} &
\underline{26.34} & \underline{3.19} & \underline{5.68} & \underline{4.23} \\

\midrule
\rowcolor{gray!10}
\textbf{MonoMSK (Ours)} &
\shortstack{\textbf{24.36} \\ \textcolor{blue}{\scriptsize (5.4\%↓)}} &
\shortstack{\textbf{1.93} \\ \textcolor{blue}{\scriptsize (32.0\%↓)}} &
\shortstack{\textbf{4.38} \\ \textcolor{blue}{\scriptsize (30.5\%↓)}} &
\shortstack{\textbf{3.17} \\ \textcolor{blue}{\scriptsize (38.8\%↓)}} &
\shortstack{\textbf{25.62} \\ \textcolor{blue}{\scriptsize (3.5\%↓)}} &
\shortstack{\textbf{2.57} \\ \textcolor{blue}{\scriptsize (18.1\%↓)}} &
\shortstack{\textbf{4.61} \\ \textcolor{blue}{\scriptsize (32.6\%↓)}} &
\shortstack{\textbf{3.33} \\ \textcolor{blue}{\scriptsize (23.8\%↓)}} &
\shortstack{\textbf{25.28} \\ \textcolor{blue}{\scriptsize (4.0\%↓)}} &
\shortstack{\textbf{2.84} \\ \textcolor{blue}{\scriptsize (11.0\%↓)}} &
\shortstack{\textbf{4.55} \\ \textcolor{blue}{\scriptsize (19.9\%↓)}} &
\shortstack{\textbf{3.29} \\ \textcolor{blue}{\scriptsize (22.2\%↓)}} \\

\bottomrule
\end{tabular}
}
\label{tab:sota_results_fitted}
\end{table*}

\textbf{Improvements to Kinematics-based Methods.} 
As shown in Table~\ref{tab:sota_results_fitted}, the proposed MonoMSK framework establishes a new benchmark for biomechanically consistent 3D human motion estimation across the BML-MoVi, BEDLAM, and OpenCap datasets. 
By combining transformer-based force–torque prediction with physics-aware motion integration through a differentiable ODE solver, 
our model achieves substantial gains in both kinematic accuracy and physical plausibility compared to existing approaches. 
On BML-MoVi,  MonoMSK reduces the joint-angle error (MAE$_{\textit{angle}}$) by 32.0\% and the bony-landmark position error (MPBLPE) by 5.4\% relative to the BioPose baseline, while decreasing the acceleration (ACCL) and velocity (VEL) errors by 30.5\% and 38.8\%, respectively. 
Similarly, on BEDLAM, our method achieves an 18.1\% improvement in MAE$_{\textit{angle}}$, a 3.5\% gain in MPBLPE, and consistent reductions in ACCL (32.6\%) and VEL (23.8\%), confirming robustness across large-scale synthetic motion data with complex dynamics. 
For OpenCap, which contains real-world motion capture sequences, MonoMSK continues to outperform BioPose, lowering MAE$_{\textit{angle}}$ by 11.0\%, MPBLPE by 4.0\%, ACCL by 19.9\%, and VEL by 22.2\%. These consistent improvements across all benchmarks highlight the effectiveness of integrating differentiable physics into transformer architectures, enabling anatomically accurate, and physically interpretable human motion estimation from monocular videos. On an NVIDIA RTX A6000, MonoMSK processes the stream at 25 FPS, corresponding to 40.0 ms/frame.


\vspace{-10pt}
\subsection{Ablation Study}
\textbf{Ablation on HMR Backbones.}
Table~\ref{tab:hmr} compares different Human Mesh Recovery (HMR) backbones used to initialize the kinematic estimates in MonoMSK. The results show that the accuracy and physical consistency of MonoMSK strongly depend on the quality of the HMR initialization. Across all datasets, performance improves consistently as the backbone transitions from HMR2.0 and DiffMesh to TokenHMR, CameraHMR, and finally MQ-HMR. 
\begin{table}[!t]
\centering
\caption{Impact of Human Mesh Recovery (HMR) backbones on kinetics estimation of MonoMSK. MAE$_\lambda$: ground-reaction forces; MAE$_\tau$: joint torques (lower is better).}
\vspace{3pt}

\setlength{\tabcolsep}{4pt} 
\renewcommand{\arraystretch}{1.2}
\tiny
\resizebox{\columnwidth}{!}{%
\begin{tabular}{l|cc|cc|cc}
\toprule
\makecell{\textbf{HMR Backbone}} &
\multicolumn{2}{c|}{\textbf{BML-MoVi}} &
\multicolumn{2}{c|}{\textbf{BEDLAM}} &
\multicolumn{2}{c}{\textbf{OpenCap}} \\
& \makecell{MAE$_{\lambda}$\\(↓)} &
  \makecell{MAE$_{\tau}$\\(↓)} &
  \makecell{MAE$_{\lambda}$\\(↓)} &
  \makecell{MAE$_{\tau}$\\(↓)} &
  \makecell{MAE$_{\lambda}$\\(↓)} &
  \makecell{MAE$_{\tau}$\\(↓)} \\
\midrule
HMR2.0 \cite{goel2023humans}& 0.0162 & 0.0584 & 0.0489 & 0.0841 & 0.0398 & 0.0726 \\
DiffMesh \cite{zheng2025diffmesh} & 0.0161 & 0.0562 & 0.0471 & 0.0833 & 0.0384 & 0.0719 \\
TokenHMR \cite{dwivedi2024tokenhmr}& 0.0150 & 0.0553 & 0.0463 & 0.0802 & 0.0379 & 0.0704 \\
CameraHMR \cite{patel2025camerahmr} & 0.0144 & 0.0528 & 0.0441 & 0.0781 & 0.0367 & 0.0689 \\
\rowcolor{gray!15}
\textbf{MQ-HMR \cite{koleini2025biopose}} &
\textbf{0.0139} & \textbf{0.0498} &
\textbf{0.0422} & \textbf{0.0748} &
\textbf{0.0351} & \textbf{0.0675} \\
\bottomrule
\end{tabular}%
}
\label{tab:hmr}
\end{table}
The MQ-HMR–based MonoMSK achieves the lowest physics-based errors, with $\mathrm{MAE}_{\lambda}=0.0139$ and $\mathrm{MAE}_{\tau}=0.0498$ on BML-MoVi, outperforming the next-best CameraHMR variant. Similar improvements are observed on BEDLAM ($\mathrm{MAE}_{\lambda}=0.0422$, $\mathrm{MAE}_{\tau}=0.0748$) and OpenCap ($\mathrm{MAE}_{\lambda}=0.0351$, $\mathrm{MAE}_{\tau}=0.0675$), demonstrating consistent gains across both synthetic and real-world motion capture datasets.
These results highlight that improved mesh-based kinematic initialization directly enhances the downstream physics-based optimization in MonoMSK, enabling more accurate force and torque estimation.

\begin{figure*}[htbp] 
    \centering
    \includegraphics[width=1.0\linewidth]{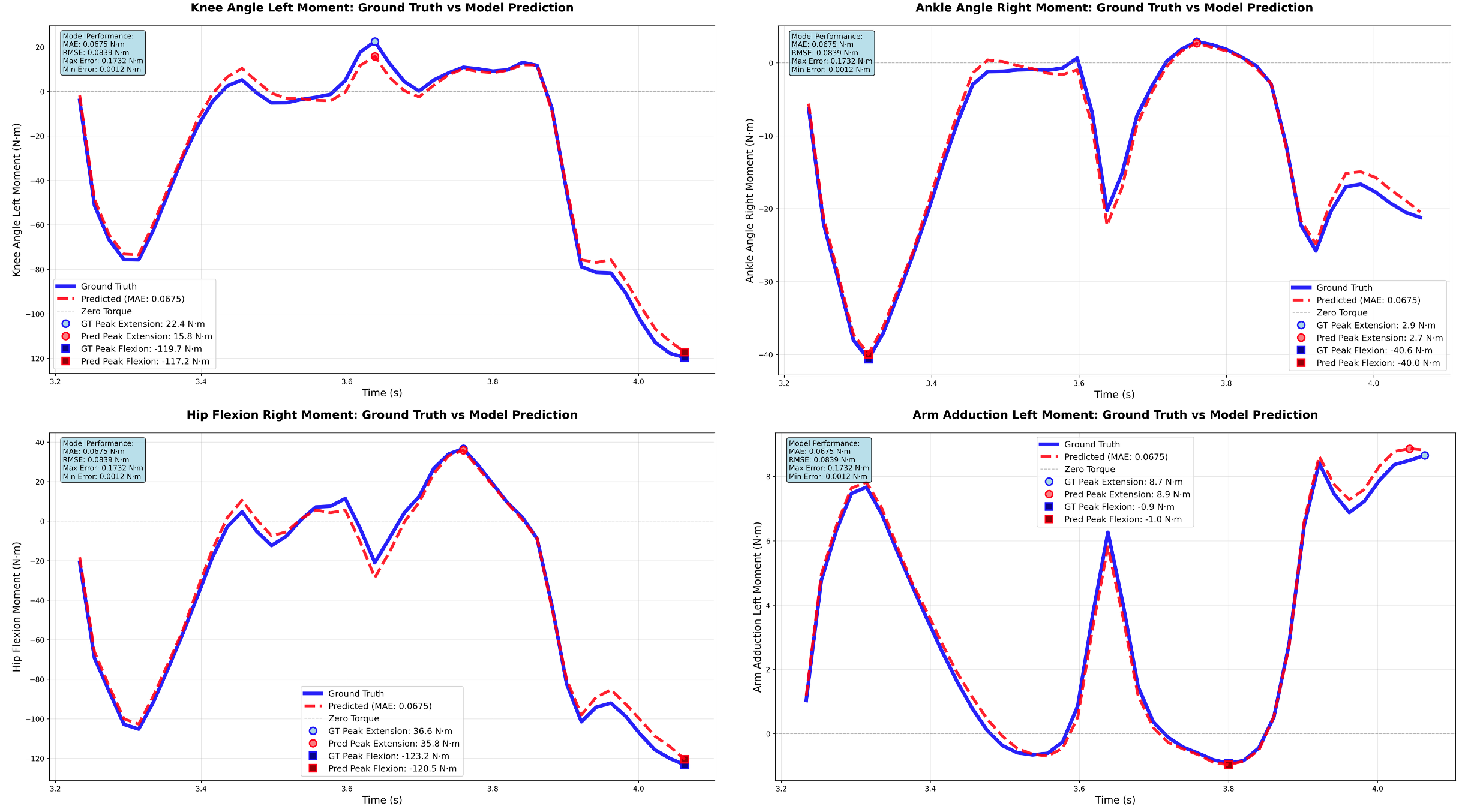}
\caption{Visualizing the estimated joint torques by MonoMSK compared with ground truth for the action sitting and standing.}
\label{fig:Joint torques}
\end{figure*}
\noindent 

\noindent\textbf{Ablation on the Training Losses.}
Table~\ref{tab:ablation_losses_MonoMSK} evaluates the contribution of each loss term. The baseline model yields the largest errors (MAE$_{\textit{angle}}$ = 2.84$^{\circ}$, MPBLPE = 25.76 mm). Adding the external force loss $\mathcal{L}_{\lambda}$ enables force estimation (MAE$_{\lambda}$ = 0.0156), while including the torque loss $\mathcal{L}_{\tau}$ further improves torque prediction (MAE$_{\tau}$ = 0.0512). Incorporating the rotation loss $\mathcal{L}_{q}$ reduces the joint angle error to 2.17$^{\circ}$. Using all losses achieves the best overall performance (MAE$_{\lambda}$ = 0.0139, MAE$_{\tau}$ = 0.0498, MAE$_{\textit{angle}}$ = 1.93$^{\circ}$, MPBLPE = 24.36 mm). Figure~\ref{fig:Joint torques} compares joint torques predicted by MonoMSK with ground-truth biomechanical simulations during a sit-to-stand task. Across all joints (knee, ankle, hip, and arm), MonoMSK closely matches both the temporal patterns and magnitudes of the torque curves, accurately capturing peak flexion/extension loads with minimal deviation (e.g., <1~N·m at the knee). The model maintains precise timing of torque reversals and low instantaneous errors (MAE $\approx$ 0.1 N·m), demonstrating its ability to infer physically meaningful joint loads directly from monocular video.

\begin{wraptable}{r}{0.48\columnwidth}  
\centering
\caption{Ablation on the Training Losses for the MonoMSK framework. Lower values (↓) indicate better performance. The first row corresponds to the kinematic-only baseline.}
\resizebox{\linewidth}{!}{%
\begin{tabular}{cccc|cccc}
\toprule
\multicolumn{4}{c|}{\textbf{Training Losses}} & \multicolumn{4}{c}{\textbf{Performance (↓)}} \\
$\mathcal{L}_{\lambda}$ & $\mathcal{L}_{\tau}$ & $\mathcal{L}_{q}$ & $\mathcal{L}_{J}$ &
MAE$_{\lambda}$ & MAE$_{\tau}$ & MAE$_{\textit{angle}}$ & MPBLPE \\
\midrule
-- & -- & \checkmark & \checkmark & -- & -- & 2.84 & 25.76 \\
\checkmark & -- & -- & -- & 0.0156 & -- & -- & -- \\
\checkmark & \checkmark & -- & -- & 0.0148 & 0.0512 & -- & -- \\
\checkmark & \checkmark & \checkmark & -- & 0.0142 & 0.0501 & 2.17 & -- \\
\checkmark & \checkmark & \checkmark & \checkmark &
\textbf{0.0139} & \textbf{0.0498} & \textbf{1.93} & \textbf{24.36} \\
\bottomrule
\end{tabular}%
}
\label{tab:ablation_losses_MonoMSK}
\end{wraptable}
\vspace{-15pt}
\newpage
\section{Conclusion}
\label{sec:conclusion}
\vspace{-5pt}
We introduce MonoMSK, the first hybrid framework to recover full-body, bio-
mechanically-accurate motion dynamics (kinematics and kinetics) from a monocular video. MonoMSK integrates learning-based inverse dynamics transformers with a differentiable, anatomically-accurate musculoskeletal (MSK) forward simulator. Data-driven inverse transformers infer kinetic causes (torques) from observed kinematic consequences (motion). Differentiable Forward Kinematics and Dynamics (FK and FD) layers act as a physics-based verifier, simulating motion from the inferred kinetics. This physics-regulated loop embeds domain knowledge during training and inference, ensuring physically plausible estimations. We train this architecture with a novel forward-inverse consistency loss, ensuring the simulated motion faithfully reconstructs the original observation. Extensive experiments on BML-MoVi, BEDLAM, and OpenCap datasets demonstrate that MonoMSK not only achieves SOTA kinematic accuracy but also delivers the first precise monocular kinetics estimation.

\clearpage


\newpage
\setcounter{page}{1}
\section{Appendix}
\label{sec:rationale}
\subsection{Overview}

The appendix is organized into the following sections:

\begin{itemize}
    \item Section \ref{sec:global_traj}: Global Trajectory Estimation
    \item Section \ref{sec:Datasets}: Datasets
    \item Section \ref{sec:Evaluation_Metrics}: Evaluation Metrics
    \item Section \ref{sec:MSK_SMPL}: SMPL vs MSK model
    \item Section \ref{sec:MSK_Model_Scaling}: MSK Model Scaling
    \item Section \ref{sec:Contact}: Selection of Body Contact Points
    \item Section \ref{sec:Kinematic_input}: Impact of Kinematic Input Data
    \item Section \ref{sec:Joint_torqe_pred}: Joint Torques and Joint  Angles Estimation
    \item Section \ref{sec:Qualitative}: Qualitative Results

\end{itemize}

\subsection{Global Trajectory Estimation}
\label{sec:global_traj}
Traditional monocular 3D human reconstruction models estimate body pose in a body-centric coordinate frame, along with a root rotation that maps the reconstructed pose into the camera frame. However, for physics-based motion modeling and dynamic analysis, it is essential to express motion in the world coordinate frame, which captures not only local joint articulation but also the person’s global motion trajectory. This requires estimating the camera-to-world rotation and the body’s world-space translation over time.

\noindent In our work, we adopt the pretrained global trajectory predictor of~\cite{yuan2022glamr} to convert body-frame joint positions into per-frame world-frame translation and rotation. Given the sequence of body-frame 3D joint positions $\{J_t\}_{t=1}^{T}$, we first extract spatial--temporal motion features using a Spatial--Temporal Graph Convolutional Network (ST-GCN)~\cite{yan2018spatial}:
\[
\{h_t\}_{t=1}^{T} = \texttt{GCN}(\{J_t\}_{t=1}^{T}),
\]
where each feature $h_t \in \mathbb{R}^{256}$ encodes both the spatial configuration and the temporal evolution of skeletal motion. Because most videos involve a fixed camera viewpoint, these features are concatenated and passed to a multi-layer perceptron (MLP) to estimate a single camera-to-world rotation:
\[
R_c = \texttt{MLP}(\{h_t\}_{t=1}^{T}).
\]
We then combine this estimated rotation $R_c$ with the root rotations produced by our kinematic mesh reconstruction module to obtain the global body rotations $\{R_t\}_{t=1}^{T}$ used in the generalized coordinate system. This predictor can be naturally extended to estimate frame-wise rotations when modeling moving-camera scenarios, but in our experiments we follow the original formulation and assume a fixed camera.

\noindent To estimate global translation, we again follow the global trajectory predictor of~\cite{carreira2016human} and employ an Iterative Error Feedback (IEF) regression module with three refinement iterations. At each frame, the IEF head takes the encoded temporal feature $h_t$ and outputs:
\[
\delta_{x,t},\; \delta_{y,t},\; z_t = \texttt{IEF}(h_t),
\]
where $(\delta_{x,t},\delta_{y,t})$ represent frame-to-frame horizontal displacement and $z_t$ denotes the vertical root position relative to the ground plane. Accumulating the horizontal displacements over time and combining them with the predicted vertical height yields the world-frame translation trajectory:
\[
\{T_t\}_{t=1}^{T}.
\]

\noindent MonoMSK incorporates this pretrained global trajectory predictor to transform body-frame joint estimates into full world-frame motion trajectories.

\subsection{Datasts}
\label{sec:Datasets}

We conduct experiments using three benchmark datasets: BML-MoVi~\cite{ghorbani2021movi}, OpenCap~\cite{uhlrich2023opencap}, and BEDLAM~\cite{black2023bedlam}. 
Our model is trained on BML-MoVi and evaluated on OpenCap and BEDLAM, enabling a thorough assessment of cross-domain generalization across real laboratory motion capture, multi-view RGB recordings, and high-fidelity synthetic motion sequences.

\vspace{4pt}
\noindent\textbf{BML-MoVi.}
BML-MoVi is a large-scale biomechanics-oriented dataset containing 90 subjects performing 21 diverse actions, including walking, running, jumping, balancing, and athletic maneuvers. 
All sequences are captured in a controlled laboratory environment using \emph{two synchronized RGB cameras} and a \emph{marker-based Vicon motion capture system}, providing accurate 3D marker trajectories.
However, BML-MoVi does not include joint-angle or musculoskeletal kinematics annotations. 
To obtain these labels, we process the raw marker trajectories using the \texttt{OpenSim Scale} and \texttt{OpenSim IK} tools.  
The Scale tool adapts a generic musculoskeletal model to each participant by estimating subject-specific bone lengths, joint centers, and inertial properties; the IK tool computes anatomically consistent joint angles by minimizing marker reconstruction error over time.
This produces high-fidelity ground-truth joint kinematics and segment orientations, making BML-MoVi suitable for supervised training of physics-informed motion estimation models.

\vspace{4pt}
\noindent\textbf{OpenCap.}
The OpenCap dataset consists of real-world motion recordings collected using \emph{five calibrated RGB cameras} in conjunction with a marker-based motion capture system.
It includes ten subjects performing functional movements such as level walking, squatting, sit-to-stand motions, drop jumps, and their asymmetric variations.
In addition to raw videos, OpenCap provides processed 3D marker trajectories, full-body joint kinematics, and a subject-specific OpenSim skeletal model.
Unlike strictly controlled laboratory datasets, OpenCap captures more natural variations in lighting, background, and viewpoint, reflecting the challenges of in-the-wild human motion capture.
This makes OpenCap an important benchmark for evaluating the robustness and real-world applicability of our physics-based motion estimation framework.

\vspace{4pt}
\noindent\textbf{BEDLAM.}
BEDLAM is a large-scale synthetic dataset containing photorealistic monocular RGB videos paired with ground-truth 3D SMPL-X meshes.
It consists of 271 subjects (109 male, 162 female) spanning a wide range of body shapes, demographics, skin tones, clothing types, hairstyles, and accessories.
The animations are produced using physically realistic simulation engines that generate dynamic body motion, detailed cloth behavior, and naturalistic secondary motion.
To integrate BEDLAM into our musculoskeletal pipeline, we convert its dense SMPL-X vertex trajectories into virtual markers and process them using \texttt{OpenSim Scale} and \texttt{OpenSim IK} to obtain musculoskeletal joint angles and body-segment parameters.
This procedure yields clean, noise-free ground-truth kinematic annotations, allowing us to evaluate model performance in a controlled synthetic setting where the underlying motion is precisely known.

\subsection{Evaluation Metrics}
\label{sec:Evaluation_Metrics}

To rigorously assess the biomechanical accuracy of MonoMSK, we employ a suite of kinematic, dynamic, and kinetic evaluation metrics. Each metric measures a different aspect of human motion quality, ranging from joint-level spatial precision to physical consistency and force reconstruction. Below, we provide detailed definitions and motivations for each metric.

\noindent\textbf{Mean Per Bony-Landmark Position Error (MPBLPE):} MPBLPE measures the 3D positional accuracy of predicted bony landmarks relative to ground-truth anatomical markers.  
This metric is inspired by the commonly used Mean Per Joint Position Error (MPJPE) in 3D pose estimation, but is adapted to the musculoskeletal context by evaluating physically meaningful anatomical landmarks. After aligning predicted and ground-truth poses at a common root (typically the pelvis), the average Euclidean distance across all bony landmarks is computed as:
\[
\text{MPBLPE}(\text{pred}, \text{target}) = \frac{1}{N} \sum_{i=1}^{N} \sqrt{\sum_{j=1}^{3} (\text{pred}_{ij} - \text{target}_{ij})^2}
\]
where \( i \) indexes the joints, and \( j \) indexes the spatial dimensions (X, Y, Z). For each joint \( i \), \( \text{pred}_{ij} \) and \( \text{target}_{ij} \) represent the predicted and target positions in dimension \( j \), respectively. The entire expression is divided by \(N\), the number of bony landmarks, to calculate the average Euclidean distance per landmark, which normalizes the loss to account for differences in the number of joints among different datasets or models. This metric reflects the fidelity of reconstructed body trajectories and segment positions, which is essential for biomechanical tasks such as movement screening, posture analysis, and joint-force computation. MPBLPE is especially important when assessing whether the predicted motion is anatomically reasonable in 3D space.

\vspace{6pt}

\noindent\textbf{Joint-Angle Mean Absolute Error (MAE$_{\textit{angle}}$):}
MAE$_{\textit{angle}}$ quantifies the mean absolute difference in joint angles (in degrees) between predicted and ground-truth rotations:
\[
\text{MAE}_{\textit{angle}} = 
\frac{1}{TK} \sum_{t=1}^{T} \sum_{j=1}^{K} 
\left| \hat{q}^{\,r}_{j,t} - q^{\,r}_{j,t} \right|.
\]

\noindent Here, \(K\) is the number of joints, and each term represents the rotational error for that joint.
In musculoskeletal simulations, **joint angles are the direct input** for the ODE-based forward dynamics and for downstream biomechanics analyses such as: joint reaction force computation, muscle activation estimation, torque estimation, and gait and clinical assessments. Errors in joint angles propagate directly into physical forces. Therefore, accurate joint-angle estimation is more critical than marker-level positional accuracy for physics-based modeling, making MAE$_{\textit{angle}}$ the primary kinematic metric in this study.

\vspace{6pt}

\noindent \textbf{Acceleration Error (ACCL):}
ACCL measures discrepancies in joint or landmark accelerations.  
Let \(\ddot{\mathbf{x}}_t\) and \(\hat{\ddot{\mathbf{x}}}_t\) denote ground-truth and predicted accelerations (computed via finite differences). The metric is:
\[
\text{ACCL} = 
\frac{1}{TM} \sum_{t=1}^{T} \sum_{i=1}^{M}
\left\| \hat{\ddot{\mathbf{x}}}^{\,i}_{t} - \ddot{\mathbf{x}}^{\,i}_{t} \right\|_2.
\]

\noindent ACCL reflects the model’s ability to produce physically smooth and dynamically consistent motion. Large acceleration deviations correspond to unrealistic “jerky” or dynamically implausible motion.

\noindent\textbf{Velocity Error (VEL):}
VEL measures discrepancies between predicted and ground-truth velocities:
\[
\text{VEL} =
\frac{1}{TM} \sum_{t=1}^{T} \sum_{i=1}^{M}
\left\| \hat{\dot{\mathbf{x}}}^{\,i}_{t} - \dot{\mathbf{x}}^{\,i}_{t} \right\|_2.
\]

\noindent VEL quantifies temporal smoothness and trajectory consistency.Accurate velocity is essential for: Realistic ,motion transitions, stable forward dynamics simulation and minimizing drift in long sequences. High VEL usually indicates noisy or jittery predictions, which can destabilize downstream physics computations.

\vspace{6pt}

\noindent \textbf{Ground-Reaction Force Error (MAE$_{\lambda}$):}
MAE$_{\lambda}$ evaluates how accurately the model predicts ground-reaction forces (GRFs):
\[
\text{MAE}_{\lambda} = 
\frac{1}{T} \sum_{t=1}^{T}
\left\| \hat{\boldsymbol{\lambda}}_{t} - \boldsymbol{\lambda}_{t} \right\|_2.
\]

\noindent GRFs result from the interaction between the foot and ground and are essential for understanding: balance, gait dynamics, impact loading, and locomotion biomechanics. GRFs reflect external forces acting on the body and are one of the hardest signals to recover from video. Accurate GRF prediction indicates that the model captures realistic foot–ground interaction and whole-body dynamics.

\vspace{6pt}

\noindent \textbf{Joint Torque Error (MAE$_{\tau}$):}
MAE$_{\tau}$ measures the accuracy of internal joint torque predictions:
\[
\text{MAE}_{\tau} =
\frac{1}{T} \sum_{t=1}^{T}
\left\| \hat{\boldsymbol{\tau}}_{t} - \boldsymbol{\tau}_{t} \right\|_2.
\]

\noindent Torques represent internal actuation produced by muscle forces and passive joint structures. Joint torques are central to musculoskeletal analysis.  
Low MAE$_{\tau}$ indicates that the model infers physically valid internal dynamics, even though torques cannot be directly observed in video. Accurate torques are essential for: understanding joint loading, injury-risk analysis, muscle-force estimation, and dynamic simulation. 

\subsection{SMPL vs MSK model}
\label{sec:MSK_SMPL}
\textbf{SMPL:} The SMPL model is a differentiable, parametric representation of the human body surface \cite{loper2023smpl}. It characterizes body geometry using pose parameters $\theta \in \mathbb{R}^{24 \times 3}$ and shape parameters $\beta \in \mathbb{R}^{10}$. The pose vector $\theta = [\theta_1, \ldots, \theta_{24}]$ contains a global body orientation $\theta_1 \in \mathbb{R}^{3}$ along with 23 local joint rotations $\theta_{2:24} \in \mathbb{R}^{23 \times 3}$, where each $\theta_k$ specifies an axis–angle rotation relative to its parent in the kinematic chain. Given $(\theta, \beta)$, SMPL generates a full-body 3D mesh $M(\theta, \beta) \in \mathbb{R}^{3 \times N}$ with $N = 6890$ surface vertices. Joint locations $J \in \mathbb{R}^{3 \times k}$ are computed via linear blend skinning, where joints are expressed as weighted combinations of mesh vertices using a predefined weight matrix $W \in \mathbb{R}^{N \times k}$ such that $J = MW$.

\begin{wrapfigure}{r}{0.5\columnwidth} 
\vspace{-10pt}
\centering
\includegraphics[width=\linewidth]{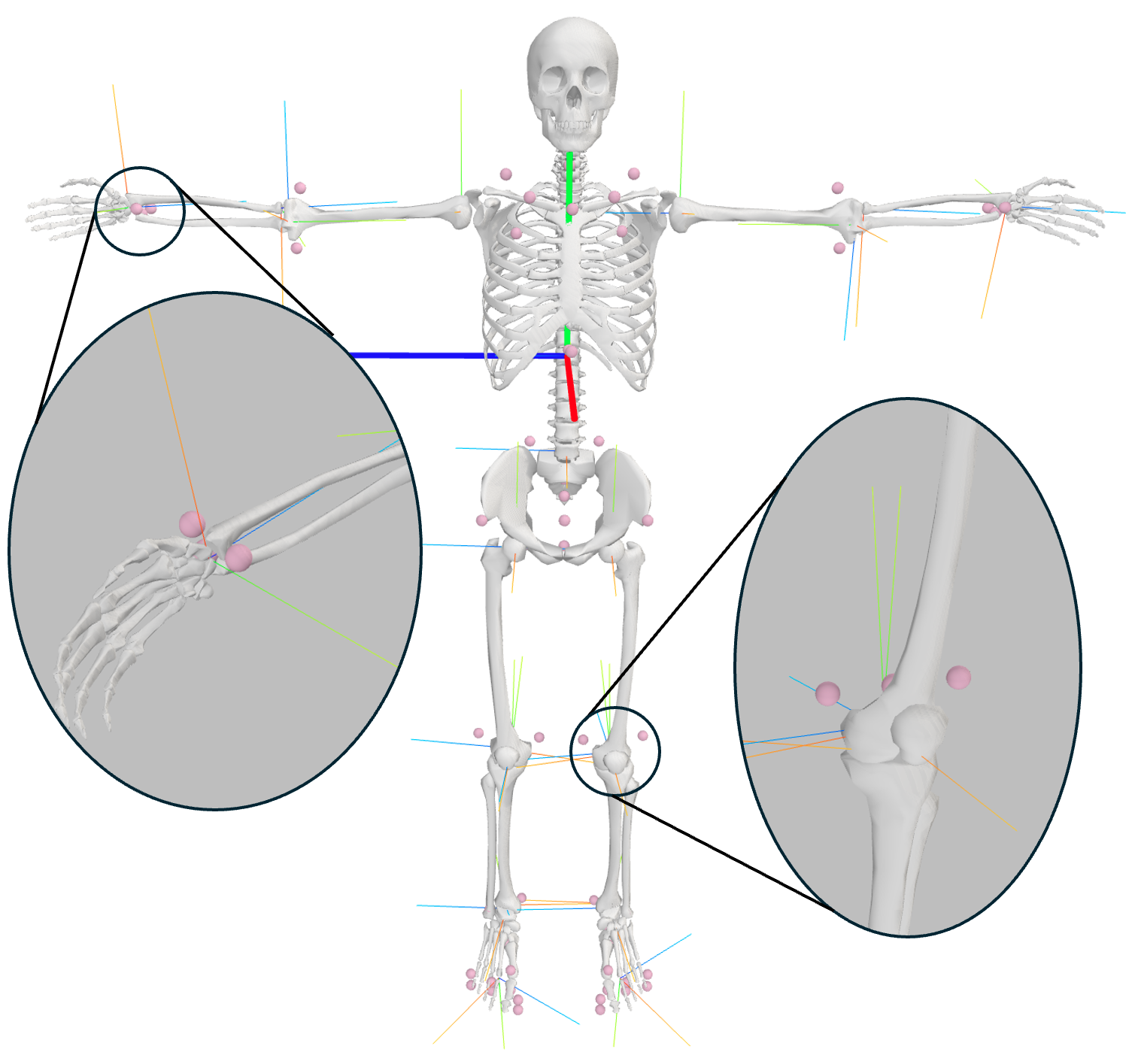}
\caption{Anatomical bone orientations. Red (X-axis): Forward, Green (Y-axis): Upward, Blue (Z-axis): Right.}
\label{fig:Bone}
\vspace{-10pt}
\end{wrapfigure}

\noindent \textbf{MSK:} The musculoskeletal (MSK) body model, commonly instantiated through platforms such as OpenSim~\cite{seth2018opensim}, represents the human body as an articulated system of $N_s = 24$ rigid bone segments connected through anatomically constrained joints. Each joint $i$ is defined by its physiological Degrees of Freedom (DoFs) $D_i$, which specify the allowable rotational axes and ultimately determine the kinematic and dynamic configuration space of the human body. 

\noindent In pharmacology, every joint is parameterized by three key components:  
(1) the anatomical bone orientation $q_i^{o} \in \mathbb{R}^{3}$, which encodes the rest-pose orientation of the bone segment (See Figure~\ref{fig:Bone});

\noindent (2) the muscle-driven joint rotation $q_i^{r} \in \mathbb{R}^{D_i}$, where $D_i \leq 3$ corresponds to the physiologically valid rotational DoFs for that joint (e.g., flexion–extension, ab/adduction, internal/external rotation); and  
(3) the bone scaling parameter $s_i \in \mathbb{R}^{3}$, which adjusts segment dimensions to match individual bone lengths and muscle attachment geometry~\cite{featherstone2008rigid,murray2017mathematical,seth2018opensim,rajagopal2016full}.  

\noindent Together, the full set of pose and scale parameters $(\mathbf{q}^{o}, \mathbf{q}^{r}, s)$ defines the complete skeletal morphology and joint configuration of the MSK model~\cite{delp2007opensim}. The MSK framework allows each subject's anatomical model to be personalized through a scaling step that aligns a generic musculoskeletal template to motion-capture or marker-based data, ensuring accurate bone lengths, segment masses, and joint center locations.

\begin{wrapfigure}{r}{0.6\columnwidth} 
\vspace{-10pt}
\centering
\includegraphics[width=\linewidth]{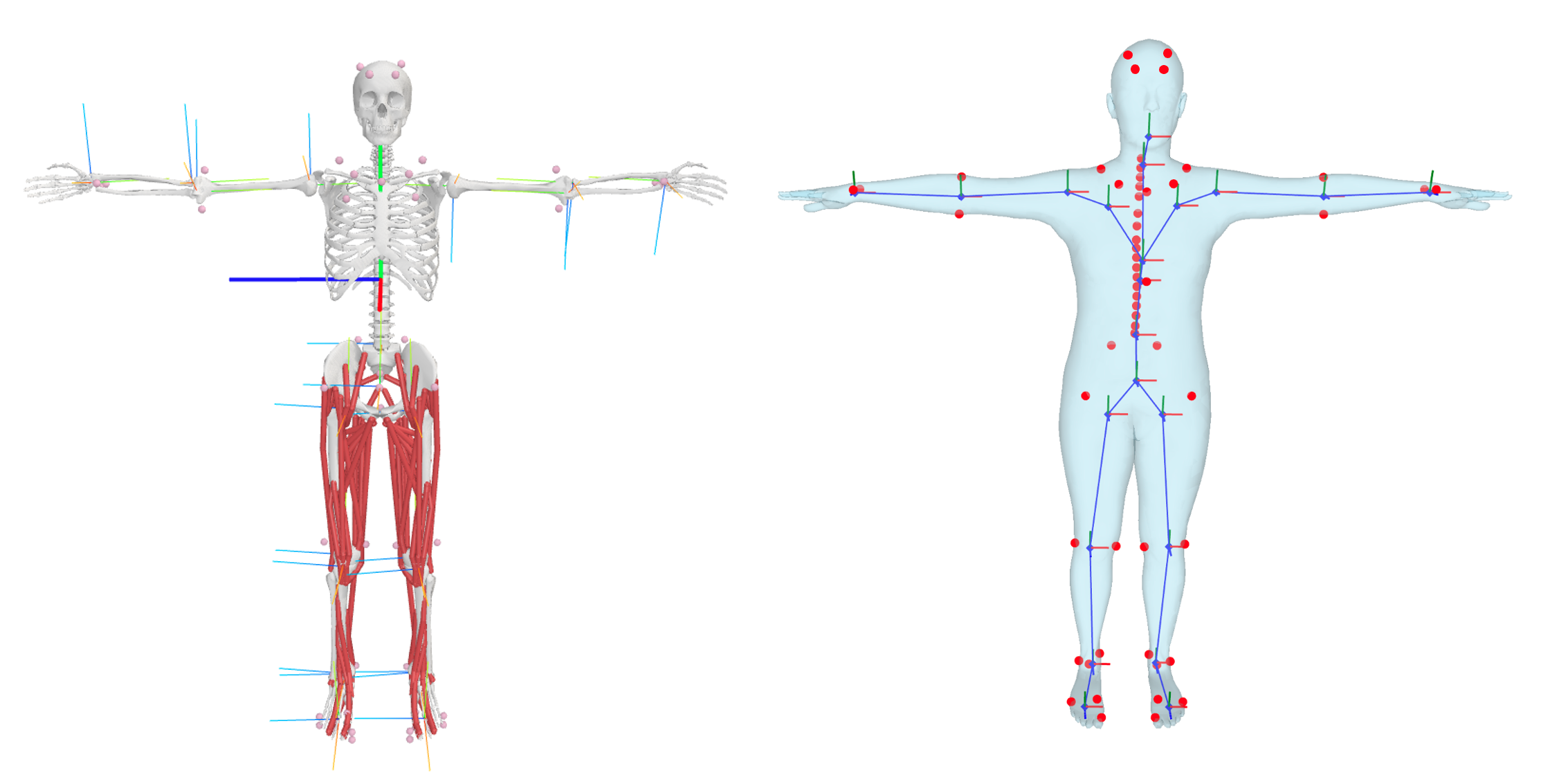}
\caption{\textbf{Left}: Musculoskeletal model has anatomical details with accurate joint locations and bone orientations and muscle geometry. \textbf{Right}: SMPL body model has a deformable 3D body surface that includes an approximate skeleton geometry with inaccurate joint locations and bone orientations.}
\label{fig:SMPL_MSK}
\vspace{-10pt}
\end{wrapfigure}

\noindent The SMPL model and MSK models fundamentally differ in purpose, structure, and the type of information they represent. SMPL is a statistical, surface-based parametric body model designed for computer vision and graphics; it encodes human shape and pose through low-dimensional parameters and produces a smooth 3D mesh with approximate joint locations defined by linear blend skinning. While highly effective for reconstructing visually plausible body geometry, SMPL lacks anatomical fidelity, its joints are not physiologically constrained, its skeleton does not correspond to real bone structures, and it cannot represent or predict physical quantities such as forces or torques. In contrast, MSK models (e.g., OpenSim) provide a biomechanically grounded representation of the human body, consisting of anatomically accurate bone segments, physiologically realistic joint degrees of freedom, subject-specific bone scaling, and detailed muscle--tendon units. These models are governed by multibody dynamics and can simulate or infer internal joint torques, muscle forces, and ground-reaction forces through forward or inverse dynamics. Whereas SMPL ``looks'' like a human body, an MSK model ``functions'' like one, enabling physically coherent motion analysis, clinical interpretation, and dynamic simulation (See Figure~\ref{fig:SMPL_MSK}).

\vspace{-10pt}
\subsection{MSK Model Scaling}
\label{sec:MSK_Model_Scaling}

With assistance of the MSK model, the kinematic analysis aims to find the optimal pose, i.e., joint rotation angles $q^{r}$, which can best fit the MSK model to the motion capture sequences.
\noindent Towards this goal, a set of model markers is first attached to the bone segments in such a way that each bone segment is associated with at least $D_i$ markers to ensure the unique solutions of the derived rotation angles at joint $i$ with $D_i$ degrees of freedom. 

\begin{wrapfigure}{r}{0.6\columnwidth} 
\vspace{-10pt}
\centering
\includegraphics[width=\linewidth]{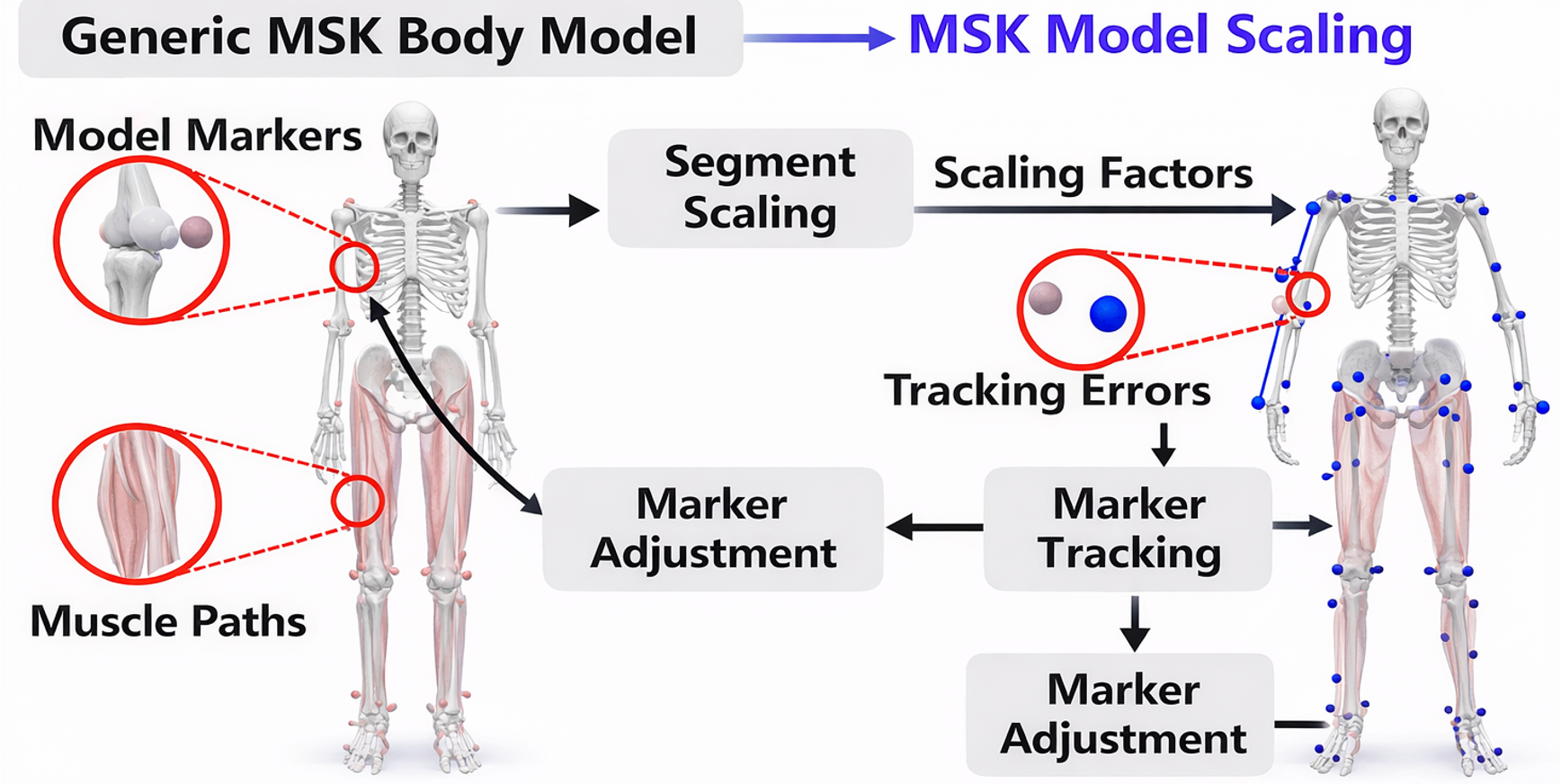}
\caption{MSK model scaling processing.}
\label{fig:MSK_scaling}
\vspace{-10pt}
\end{wrapfigure}

\noindent Then, a set of corresponding experiential markers is placed on the human subject. Then, the pose $q^{r}$ and bone scale $s$ can be obtained by solving an optimization problem that minimizes the distance between each experimental marker and its corresponding model marker. i.e., 
\begin{align*}
 q^{r*}, s^* =\arg\min_{ q^{r}, s}\sum_{i = 1}^{M} \left\| {f_{FK}(q^{r}, s, q^{o}, p_i)} - x_i^{exp}\right\|      
\end{align*}
where $M$ is the number of markers. $p_i \in \mathbb{R}^{3}$ denotes the position of $i$-th model marker in the local coordination system of the body segment to which it is attached.  $f_{FK}(q^{r}, s, q^{o}, p_i)$ is the forward kinematics transformation that converts the model marker $i$ from its local coordination frame to the world coordination system under the scaled skeleton with the pose of $q^{r}$. $x_i^{exp} \in \mathbb{R}^{3}$ is the position of the experiential marker $i$ in the world coordination system (See Figure~\ref{fig:MSK_scaling}).

\begin{wrapfigure}{r}{0.6\columnwidth} 
\vspace{-10pt}
\centering
\includegraphics[width=\linewidth]{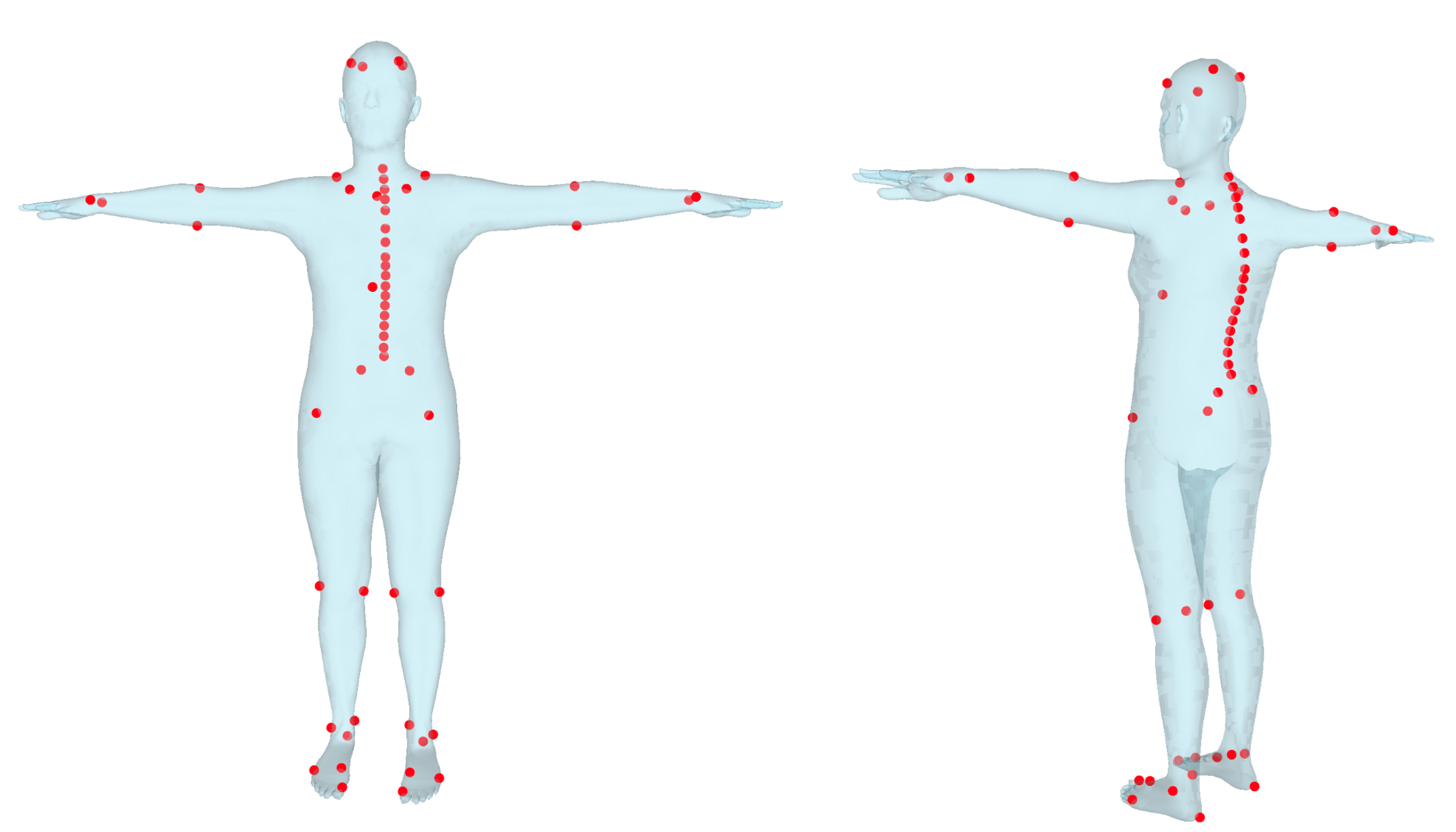}
\caption{The markers defined on SMPL (front and back view).}
\label{fig:markerset}
\vspace{-10pt}
\end{wrapfigure}

\noindent To fit MSK to SMPL, we define the same markers on both models. Theoretically, we could define each skin vertex of SMPL to be a marker attached
to BSM. Specifically, we define 52 bony markers that are close to the bones, as typically done in motion capture. Each marker is defined on MSK and SMPL by identifying specific SMPL vertices. Figure~\ref{fig:markerset}) shows all the bony markers on SMPL.

\vspace{-10pt}
\subsection{Contact Points}
\label{sec:Contact}

Accurate modeling of foot--ground interaction is essential for 
estimating physically meaningful ground reaction forces (GRFs) 
and joint kinetics in musculoskeletal simulation. 
Following established OpenSim conventions 
\cite{rajagopal2016full,seth2018opensim}, as well as the 
OpenCap framework \cite{uhlrich2023opencap}, we adopt a 
twelve-element compliant contact model consisting of six 
spherical contact points per foot. 
This model provides stable, high-fidelity approximations of 
plantar pressure distribution and GRF trajectories without 
requiring instrumented force plates.

\paragraph{Anatomical Placement of Contact Spheres.}
Each foot is equipped with six contact spheres positioned at 
biomechanically meaningful locations:  
\begin{itemize}
    \item two spheres on the posterior--lateral and 
          posterior--medial calcaneus (heel strike region),  
    \item two spheres on the medial and lateral midfoot 
          (arch support region),  
    \item two spheres on the first and fifth metatarsal heads 
          (forefoot and toe-off region).  
\end{itemize}
These locations capture the full progression of the stance 
phase, including heel strike, mid-stance load distribution, 
and toe-off propulsion (See Figure~\ref{fig:Contact}). The same arrangement is used in the 
Rajagopal full-body musculoskeletal model 
\cite{rajagopal2016full} and subsequently adopted by OpenCap 
\cite{uhlrich2023opencap}.

\paragraph{Normal Contact Force.}
Foot--ground contact forces are computed using the nonlinear 
Hunt--Crossley compliant contact formulation 
\cite{hunt1975coefficient}. 

\begin{wrapfigure}{r}{0.38\columnwidth} 
\vspace{-10pt}
\centering
\includegraphics[width=\linewidth]{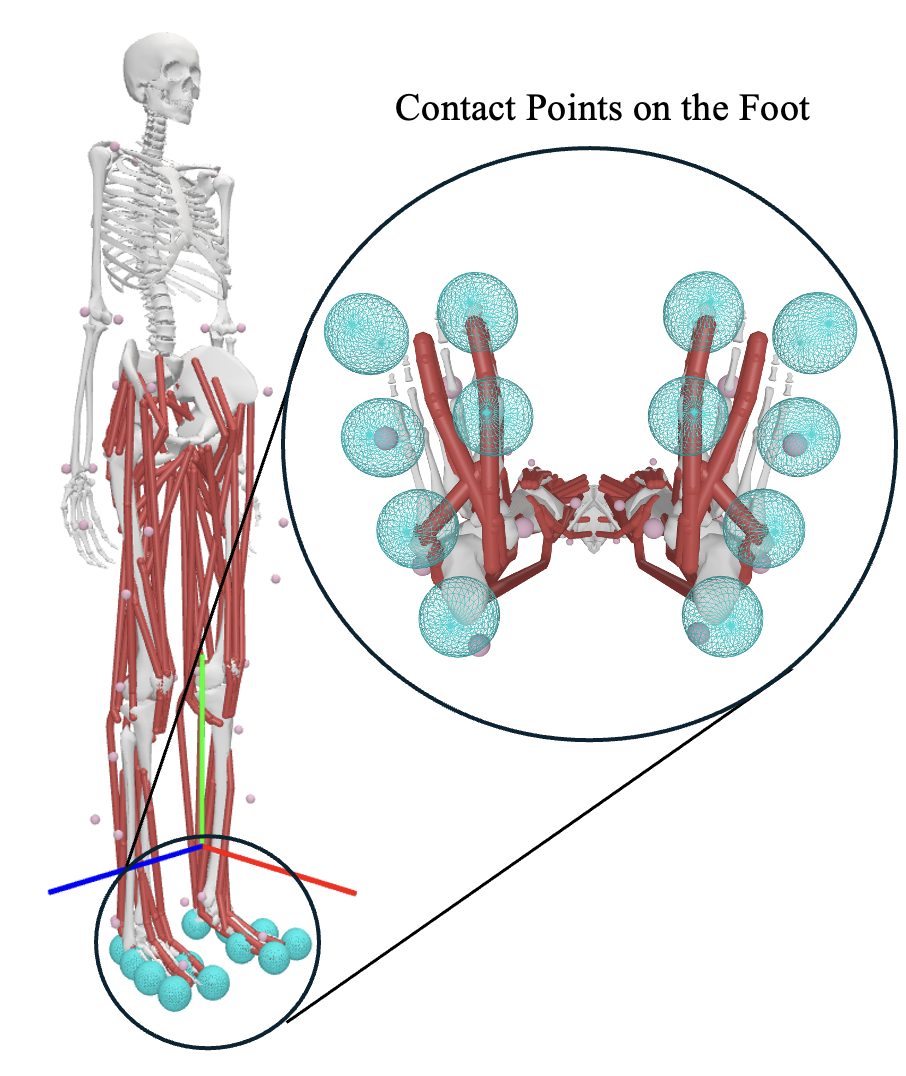}
\caption{Foot–ground contact model with six contact spheres per foot (12 total).}
\label{fig:Contact}
\vspace{-10pt}
\end{wrapfigure}
\noindent For a given contact sphere $k$, the normal force 
$\mathbf{F}_{n,k}$ is defined as:
\begin{equation}
\mathbf{F}_{n,k}
= k_n \, \delta_k^{1.5} 
  \bigl( 1 + c_n \dot{\delta}_k \bigr) \, \mathbf{n},
\label{eq:hc_normal}
\end{equation}
where $\delta_k$ is the penetration depth between the sphere 
and ground, $\dot{\delta}_k$ is the penetration velocity, 
$\mathbf{n}$ is the ground normal vector, 
$k_n$ is the contact stiffness constant, and 
$c_n$ is the damping coefficient. 
The $\delta^{1.5}$ term provides nonlinear stiffness and 
ensures smooth force generation during heel strike and 
toe-off.

\paragraph{Tangential Friction Force.}
The tangential (shear) component of the GRF is modeled using a 
smooth Coulomb-like friction law:
\begin{equation}
\mathbf{F}_{t,k}
=
-\mu \, \|\mathbf{F}_{n,k}\| 
\frac{\mathbf{v}_{t,k}}
     {\|\mathbf{v}_{t,k}\| + \epsilon},
\label{eq:hc_tangential}
\end{equation}
where $\mu$ is the coefficient of friction, 
$\mathbf{v}_{t,k}$ is the tangential slip velocity at the 
contact point, and $\epsilon$ is a small regularization 
constant ensuring numerical stability. 
This formulation yields realistic shear behavior during foot 
roll-over, directional changes, and push-off.

\paragraph{Resultant Ground Reaction Force.}
The total GRF acting on each foot at time $t$ is the sum of 
forces produced by its six contact spheres:
\begin{equation}
\boldsymbol{\lambda}(t)
=
\sum_{k=1}^{6}
\left( \mathbf{F}_{n,k}(t) + \mathbf{F}_{t,k}(t) \right).
\label{eq:grf_total}
\end{equation}
This aggregated GRF serves as the external force input to the 
Euler--Lagrange dynamics, 
and is critical for estimating internal joint torques and 
muscle-driven motion.

\paragraph{Advantages of the 12-Point Contact Model.}
The twelve-sphere foot--ground interaction model:
\begin{itemize}
    \item captures heel strike, flat-foot, and toe-off phases,  
    \item enables realistic medial--lateral GRF distribution,  
    \item avoids discontinuities associated with rigid contact,  
    \item is differentiable and compatible with learning-based 
          pipelines,  
    \item requires no force plates or pressure insoles.  
\end{itemize}
These properties make the model widely adopted for simulations 
in biomechanics, clinical gait analysis, and data-driven 
human motion reconstruction.

\vspace{-10pt}
\subsection{Impact of Kinematic Input Data.} 
\label{sec:Kinematic_input}
\begin{table*}[!t]
\centering
\caption{Ablation on kinematic input source. Comparison between kinematics predicted pose data and ground-truth (GT) kinematics for training and testing on BML, BEDLAM, and OpenCap datasets. Lower values ($\downarrow$) indicate better performance.}
\vspace{3pt}
\scalebox{0.9}{
\begin{tabular}{l|ccc|ccc}
\toprule
 & \multicolumn{3}{c|}{Kinematics Predicted Pose Data} & \multicolumn{3}{c}{GT Pose Data} \\
\textbf{Metrics} & \textbf{BML} & \textbf{OpenCap} & \textbf{BEDLAM} & \textbf{BML} & \textbf{OpenCap} & \textbf{BEDLAM} \\
\midrule
MAE$_{\lambda}$ & 0.0138 & 0.0353 & 0.0423 & 0.0112 & 0.0215 & 0.0327 \\
MAE$_{\tau}$ & 0.0499 & 0.0671 & 0.0749 & 0.0399 & 0.0472 & 0.0519 \\
MAE$_{\textit{kinetics}}$ & 0.0632 & 0.0712 & 0.0823 & 0.0231 & 0.0513 & 0.0614 \\
\bottomrule
\end{tabular}
}
\label{tab:kinematics_ablation}
\end{table*}

To evaluate the influence of input kinematic quality on the physics-based performance of MonoMSK, we conduct an ablation comparing two input configurations: (1) \textit{Predicted Kinematics}, where the motion inputs $\{\hat{\mathbf{q}}_t\}$ are obtained from the BioPose model, and (2) \textit{Ground-Truth (GT) Kinematics}, where true joint trajectories $\{\mathbf{q}_t^{\mathrm{GT}}\}$ are used. 
As shown in Table~\ref{tab:kinematics_ablation}, the model using GT kinematics consistently achieves lower physics-based errors across all datasets. 
Specifically, the force loss MAE$_{\lambda}$ decreases from 0.0138 to 0.0112 on BML, 0.0353 to 0.0215 on OpenCap, and 0.0423 to 0.0327 on BEDLAM, corresponding to improvements of 19.0\%, 39.1\%, and 22.7\%, respectively. 
Similarly, the joint torque loss MAE$_{\tau}$ drops by 20–30\% across datasets, while the overall kinetics loss MAE$_{\textit{kinetics}}$ is reduced by up to 40\%. 
These results highlight the strong dependency of accurate force–torque prediction on high-quality kinematic inputs, confirming that MonoMSK effectively propagates improvements in motion estimation to enhanced physical fidelity. 
This demonstrates the framework’s ability to translate precise kinematic representations into physically consistent dynamics, even under cross-dataset conditions.

\begin{table*}[!t]
\centering
\caption{Comparison between Single Frame Out and Multiple Frames Out MonoMSK models across three datasets. Lower values ($\downarrow$) indicate better performance.}
\vspace{3pt}
\scalebox{0.9}{
\begin{tabular}{l|ccc|ccc|ccc}
\toprule
 & \multicolumn{3}{c|}{\textbf{BML-MoVi}} & \multicolumn{3}{c|}{\textbf{BEDLAM}} & \multicolumn{3}{c}{\textbf{OpenCap}} \\
\textbf{Model} & MAE$_{\lambda}$ & MAE$_{\tau}$ & MAE$_{\textit{angle}}$ &
MAE$_{\lambda}$ & MAE$_{\tau}$ & MAE$_{\textit{angle}}$ &
MAE$_{\lambda}$ & MAE$_{\tau}$ & MAE$_{\textit{angle}}$ \\
\midrule
Multiple Frames Out & 0.0156 & 0.0538 & 2.45 & 0.0472 & 0.0785 & 2.81 & 0.0373 & 0.0682 & 3.42 \\
\rowcolor{gray!15}
Single Frame Out & 0.0139 & 0.0498 & \textbf{1.93} & 0.0422 & 0.0748 & \textbf{2.57} & 0.0351 & 0.0675 & \textbf{2.84} \\
\bottomrule
\end{tabular}
}
\label{tab:temporal_ablation}
\end{table*}
\noindent\textbf{Single Frame Out vs. Multiple Frames Out.}
We compare two temporal strategies in MonoMSK: (1) \textit{Single Frame Out}, which predicts one frame at a time with end-to-end ODE supervision, and (2) \textit{Multiple Frames Out}, which predicts several future frames in a two-stage setup without direct physical supervision. As shown in Table~\ref{tab:temporal_ablation}, Single Frame Out consistently performs better across all datasets, achieving lower force and torque errors (MAE${\lambda}$, MAE${\tau}$) and improved joint-angle accuracy. For instance, on BML-MoVi it reduces MAE${\lambda}$ from 0.0156 to 0.0139, MAE${\tau}$ from 0.0538 to 0.0498, and MAE$_{\textit{angle}}$ from 2.45° to 1.93°. These results suggest that step-wise prediction benefits from stronger physical supervision and reduces temporal drift, making autoregressive single-step prediction more effective for physically grounded motion modeling.

\begin{figure*}[htbp] 
    \centering
    \includegraphics[width=1.0\linewidth]{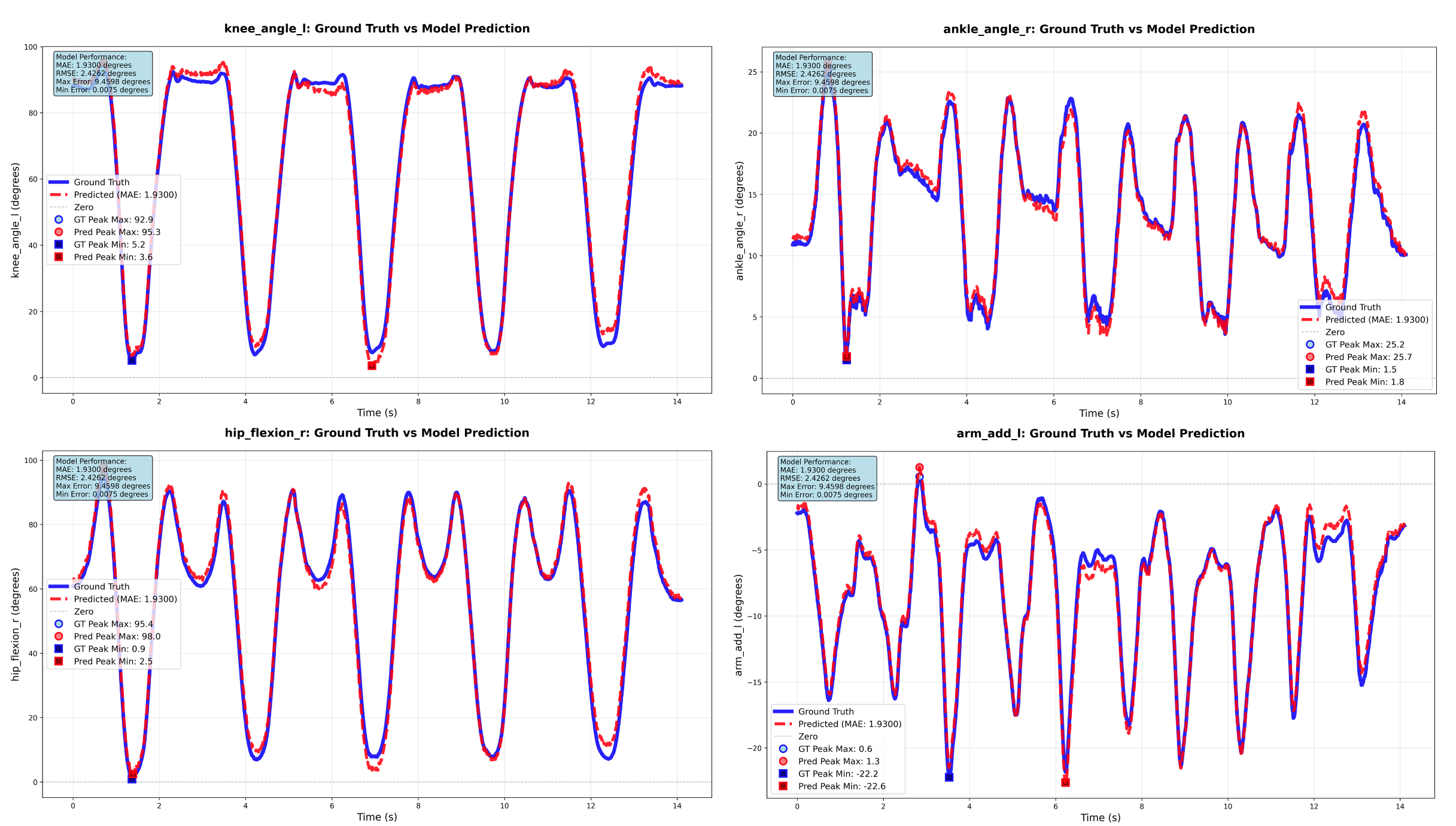}
\caption{Visualizing the estimated joint angles by MonoMSK compared with ground truth for the action sitting and standing.}
\label{fig:Joint angles}
\end{figure*}
\vspace{-10pt}
\subsection{Joint Torques and Joint Angles Estimation}
\label{sec:Joint_torqe_pred}
Figure~\ref{fig:Joint angles} presents a qualitative and quantitative comparison between MonoMSK’s predicted joint angles and ground-truth measurements across several representative joints during a sit-to-stand motion sequence. The predicted trajectories (red) closely track the ground-truth curves (blue), exhibiting minimal phase delay and small amplitude discrepancies throughout both dynamic and quasi-static phases. Across joints such as knee flexion, ankle dorsiflexion, hip flexion, and arm adduction, MonoMSK achieves consistently low mean absolute errors (MAE $\approx$ 1.9$^\circ$), accurately reproducing both peak angles and full temporal profiles. Peak-angle differences remain within 1--2$^\circ$ for most joints, confirming the model’s ability to capture rapid flexion–extension cycles and subtle postural adjustments with high fidelity. These results demonstrate that MonoMSK delivers near–ground-truth accuracy in joint-angle reconstruction, preserving anatomical coherence and temporal smoothness across the full motion sequence.

\subsection{Qualitative Results}
\label{sec:Qualitative}

Figure~\ref{fig:MonoMSK-qual} presents qualitative results of MonoMSK across a range of real-world motions, including drop jumps, sit-to-stand transitions, and seated movements. From each monocular input frame, the method reconstructs a full musculoskeletal (MSK) body model and visualizes it from multiple viewpoints. The recovered poses exhibit anatomically consistent joint configurations, realistic muscle–tendon paths, and stable global alignment, even during rapid or highly dynamic motions.

\noindent In the visualizations, pink markers denote the tracked virtual marker locations, while the blue structures represent the estimated muscle–tendon units. Across different subjects and activities, MonoMSK maintains coherent lower-body kinematics and physically plausible musculoskeletal geometry, demonstrating the model's ability to recover biomechanically meaningful motion from monocular video.

\noindent Figure~\ref{fig:Limitation} further illustrates a dynamic motion sequence over multiple timesteps, showing the reconstructed MSK body from front, side, and back views. While the method captures realistic lower-body motion and contact-aware posture changes, minor inaccuracies can appear in some upper-body regions, particularly the head and arms. This limitation arises because MonoMSK primarily focuses on modeling lower-limb biomechanics and foot–ground interactions, which provide the strongest physical constraints for musculoskeletal inference.

\begin{figure*}[htbp] 
    \centering
    \includegraphics[width=0.8\linewidth]{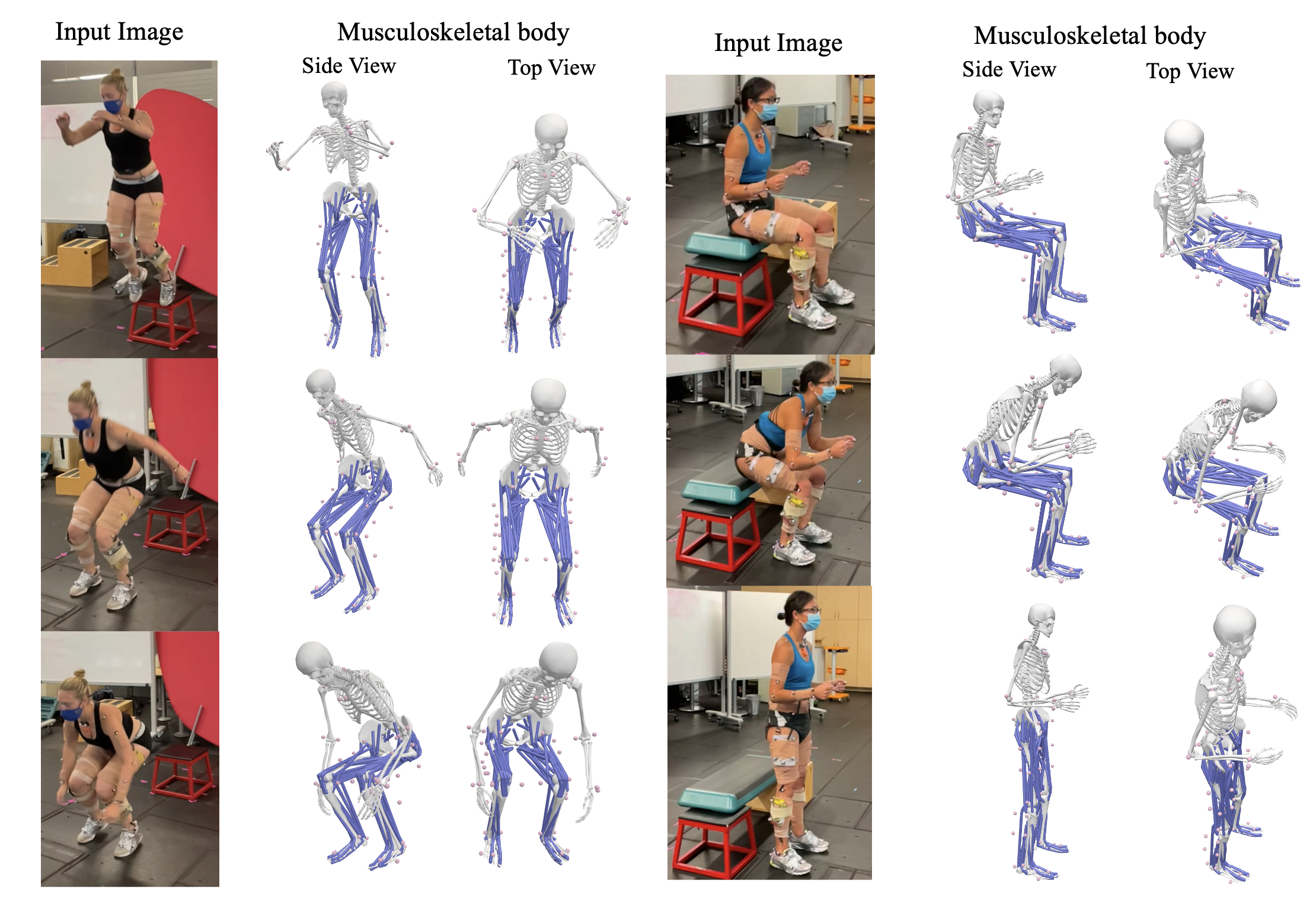}
\caption{Qualitative examples of MonoMSK on diverse motions. From monocular images, the method reconstructs a musculoskeletal body model with estimated muscle–tendon units (blue) and virtual markers (pink), shown from multiple viewpoints.}
\label{fig:MonoMSK-qual}
\end{figure*}

\begin{figure*}[htbp] 
    \centering
    \includegraphics[width=0.8\linewidth]{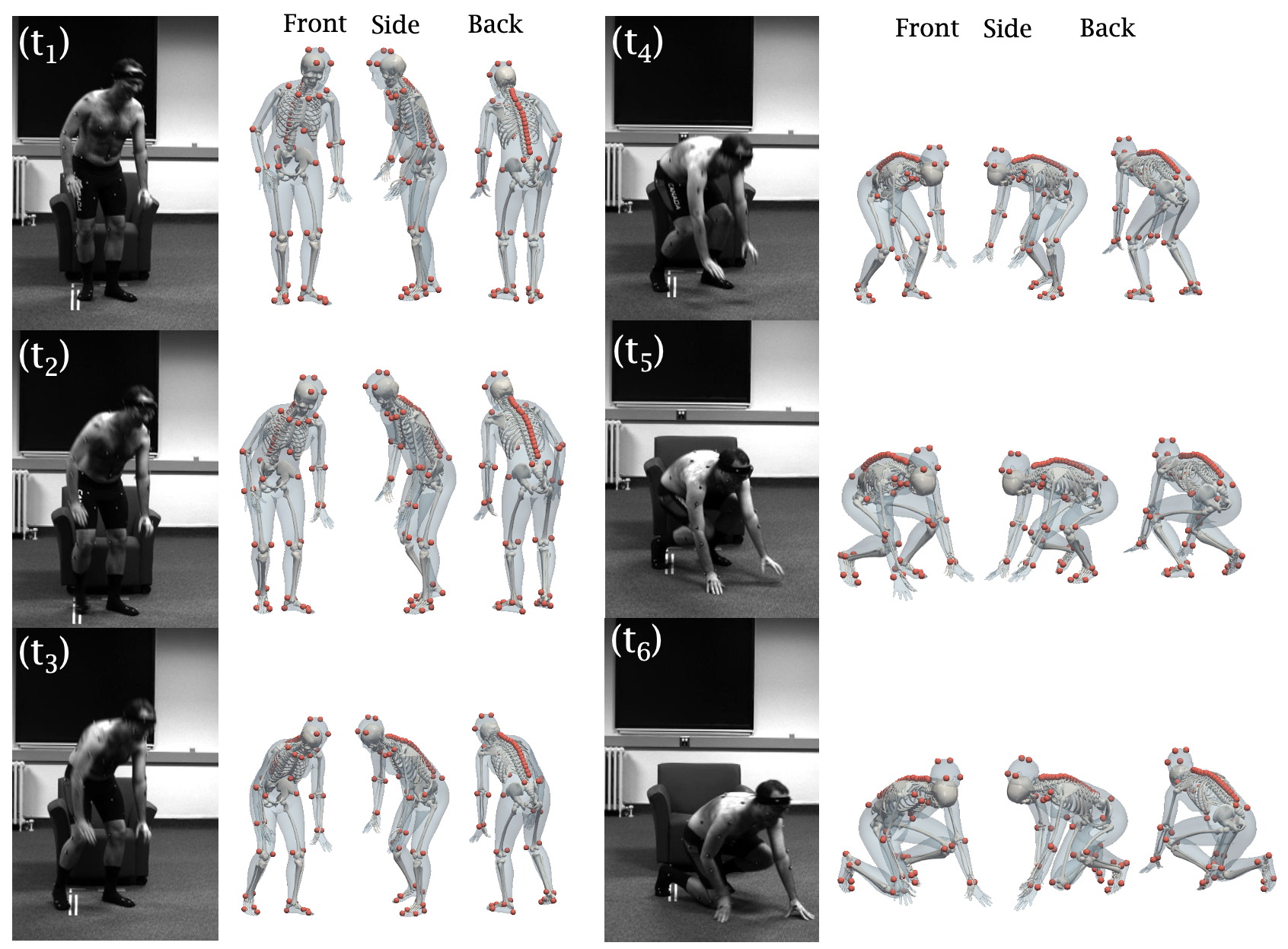}
\caption{Example sequence showing reconstructed musculoskeletal poses over time ($t_1$–$t_6$). While lower-body kinematics and contact-aware motion are well captured, some inaccuracies may appear in upper-body regions such as the head and hands.}
\label{fig:Limitation}
\end{figure*}

\newpage
\bibliographystyle{splncs04}
\bibliography{main}

\end{document}